\documentclass{article}

\usepackage{microtype}
\usepackage{graphicx}
\usepackage{booktabs} 
\usepackage{colortbl}
\definecolor{pgc}{gray}{.80}
\definecolor{bgc}{gray}{.95}
\usepackage{hyperref}
\usepackage{float}                  
\usepackage{subfig}                 
\usepackage{overpic} 
\usepackage{multirow}

\usepackage[accepted]{icml2024}

\usepackage{amsmath}
\usepackage{amssymb}
\usepackage{mathtools}
\usepackage{amsthm}
\usepackage{enumerate}

\usepackage[capitalize,noabbrev]{cleveref}

\theoremstyle{plain}
\newtheorem{theorem}{Theorem}[section]

\theoremstyle{definition}

\theoremstyle{remark}
\newtheorem{remark}[theorem]{Remark}

\usepackage[textsize=tiny]{todonotes}

\icmltitlerunning{\algo: Robust Prompt Tuning with Out-of-Distribution Detection}

\newcommand{\algo}{\textsc{DeCoOp}}
\newcommand{\coop}{\textsc{CoOp}}
\newcommand{\cocoop}{\textsc{CoCoOp}}
\newcommand{\ship}{\textsc{Ship}}
\newcommand{\Clip}{\textsc{Clip}}
\newcommand{\Pens}{\textsc{Prompt Ens.}}

\newcommand{\setting}{\text{OPT}}
\newcommand{\dept}{\textsc{DePt}}
\newcommand{\bs}{\text{b}}
\newcommand{\nw}{\text{n}}
\newcommand{\pt}{\textsc{Pt}}
\newcommand{\zs}{\textsc{Zs}}
\newcommand{\ood}{\textsc{Ood}}
\newcommand{\cls}{\textsc{Cls}}

\begin{document}

\twocolumn[
\icmltitle{\algo: Robust Prompt Tuning with Out-of-Distribution Detection}
\begin{icmlauthorlist}
\icmlauthor{Zhi Zhou}{njulab}
\icmlauthor{Ming Yang}{njulab,njuai}
\icmlauthor{Jiang-Xin Shi}{njulab,njuai}
\icmlauthor{Lan-Zhe Guo}{njulab,njuict}
\icmlauthor{Yu-Feng Li}{njulab,njuai}
\end{icmlauthorlist}
\icmlaffiliation{njulab}{National Key Laboratory for Novel Software Technology, Nanjing University, China}
\icmlaffiliation{njuict}{School of Intelligence Science and Technology, Nanjing University, China}
\icmlaffiliation{njuai}{School of Artificial Intelligence, Nanjing University, China}
\icmlcorrespondingauthor{Lan-Zhe Guo}{guolz@nju.edu.cn}
\icmlcorrespondingauthor{Yu-Feng Li}{liyf@nju.edu.cn}
\icmlkeywords{Prompt Tuning, Out-of-Distribution Detection, Vision-Language Model, Foundation Model}
\vskip 0.3in
]



\printAffiliationsAndNotice{}  

\begin{abstract}
Vision-language models (VLMs), such as CLIP, have demonstrated impressive zero-shot capabilities for various downstream tasks. Their performance can be further enhanced through few-shot prompt tuning methods. However, current studies evaluate the performance of learned prompts separately on base and new classes. This evaluation lacks practicality for real-world applications since downstream tasks cannot determine whether the data belongs to base or new classes in advance. 
In this paper, we explore a problem setting called \emph{\textbf{O}pen-world \textbf{P}rompt \textbf{T}uning} (\setting), which involves tuning prompts on base classes and evaluating on a combination of base and new classes. 
By introducing \emph{\textbf{De}composed \textbf{P}rompt \textbf{T}uning} framework (\dept), we theoretically demonstrate that \setting\ can be solved by incorporating out-of-distribution detection into prompt tuning, thereby enhancing the base-to-new discriminability. 
Based on \dept, we present a novel prompt tuning approach, namely, \emph{\textbf{De}composed \textbf{Co}ntext \textbf{Op}timization} (\algo), which introduces new-class detectors and sub-classifiers to further enhance the base-class and new-class discriminability. 
Experimental results on 11 benchmark datasets validate the effectiveness of \dept\ and demonstrate that \algo\ outperforms state-of-the-art methods, providing a significant 2\% average accuracy improvement. 
\end{abstract}

\section{Introduction}

Vision-language models (VLMs), such as CLIP~\cite{clip}, have been developed to align images and language, demonstrating impressive zero-shot capabilities for a variety of downstream tasks~\cite{imagenet, fgvc, stanfordcars}, using only class names. The classification prediction is determined by calculating the cosine similarity between the image embedding, generated by the image encoder, and the text embedding, generated by the text encoder, using prompting techniques~\cite{liu2023pre}. For example, by inputting ``a photo of {class}",  the text encoder generates the corresponding text embedding, where ``{class}" represents the class name.

\begin{figure}[t]
    \begin{center}
    \centerline{\includegraphics[width=\columnwidth]{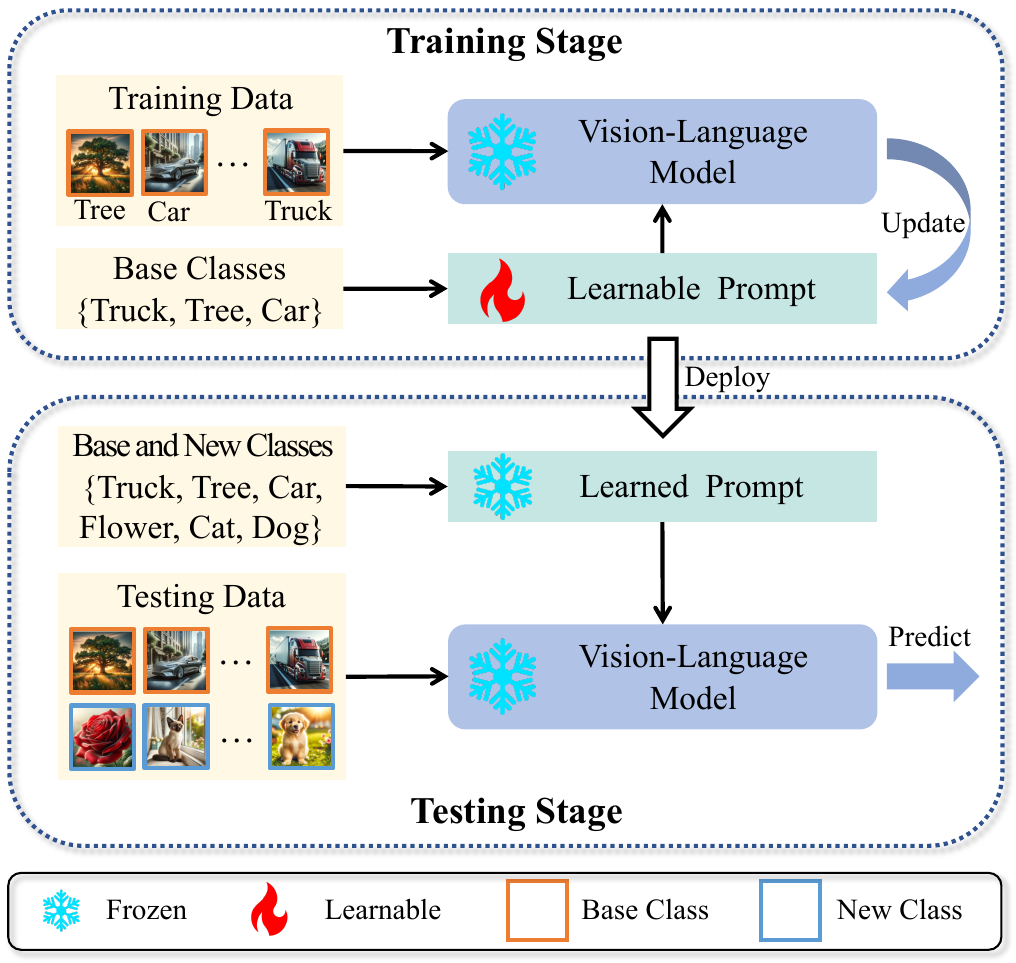}}
    \caption{An illustration of the \setting\ evaluation paradigm. During the training, we finetune the model with data from base classes. During the testing, we evaluate the model on a mix of base and new classes.}
    \label{fig: Setting}
    \end{center}
    \vskip -0.2in
\end{figure}

In addition, it is possible to improve the performance of CLIP, particularly when dealing with downstream tasks that have limited labeled data. Few-shot prompt tuning methods~\cite{proda, coop, shu2022test} utilize a small amount of labeled data from downstream datasets to fine-tune learnable prompts while keeping the other parameters unchanged. These approaches can yield substantial performance improvement compared to the zero-shot VLMs in downstream classification tasks.
However, previous studies~\cite{cocoop, ship} have identified a limitation in which the learned prompts only operate effectively with labeled data from base classes. This limitation leads to a decrease in zero-shot performance for new classes which are unseen in the training set.
To address this, the researchers propose an evaluation paradigm that assesses the performance of both base and new classes separately, as well as their harmonic mean, i.e., H metric.

\begin{figure}[t]
    \begin{center}
    \subfloat[Enhancement in H metric leads to reduced accuracy]{\includegraphics[width=\columnwidth]{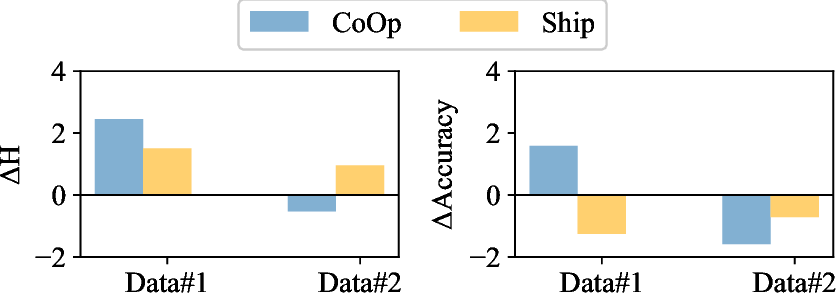}}\\
    \subfloat[Deterioration in H metric leads to improved accuracy]{\includegraphics[width=\columnwidth]{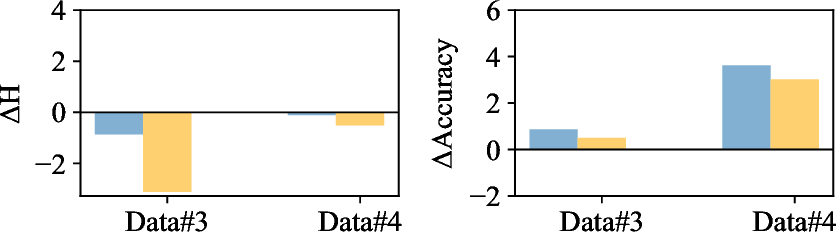}}
    \caption{Delta performance of \coop\ and \ship\ method compared to zero-shot baseline \Clip\ method. Subfigres (a) and (b) show that the changes in the H metric are not necessary indicators of performance improvements or degradation of accuracy, highlighting the significance of addressing the \setting\ problem.}
    \label{fig: DeltaAccuracy}
    \end{center}
    \vskip -0.2in
\end{figure}

Although this evaluation paradigm can comprehensively evaluate the performance of both base and new classes, it lacks practicality for real-world applications, which necessitate prior knowledge of whether the data belongs to base or new classes in the downstream task.
For instance, in the context of biological underpinnings~\cite{HayesKBSSK21, KudithipudiABBB22} and visual classification~\cite{LangeAMPJLST22, MaiLJQKS22}, both base classes and new classes that arise during testing will be evaluated together.
Therefore, we introduce a realistic problem setting, namely, \emph{\textbf{O}pen-world \textbf{P}rompt \textbf{T}uning} (\setting), which evaluates the performance of the model on a mix of base and new classes while training model with base classes.
An illustration of the \setting\ problem is shown in \autoref{fig: Setting}. 
The results in \autoref{fig: DeltaAccuracy} show that the changes in the H metric are not necessary indicators of performance improvement or degradation when evaluating the combination of base and new classes, which highlights the significance of the \setting\ problem. 

To address the \setting\ problem, we first analyze the original problem, which consists of three parts: base-to-new discriminability, base-class discriminability, and new-class discriminability. We observe that existing methods and settings fail to adequately consider the base-to-new discriminability. 
Motivated by this analysis, we propose the \dept\ framework, which incorporates out-of-distribution (OOD) detection into prompt tuning to enhance the base-to-new discriminability and thereby prevents performance degradation on new classes. We theoretically prove that the \dept\ framework can improve performance compared to the zero-shot baseline and prompt tuning methods. 
Building upon the \dept\ framework, we introduce a novel prompt tuning approach called \emph{\textbf{De}composed \textbf{Co}ntext \textbf{Op}timization} (\algo). This approach incorporates new-class detectors and sub-classifiers to further enhance the base-class and new-class discriminability, respectively. 
Empirical results validate the effectiveness of the \dept\ framework and demonstrate that \algo\ approach outperforms current state-of-the-art (SOTA) methods by a significant margin.

The contributions of this paper are summarized as follows: 
\begin{enumerate}[(1)]
    \item We explore a practical \setting\ problem and break down the problem into two sub-problems: OOD detection and prompt tuning. Through decomposition, we uncover that base-to-new discriminability is crucial to address \setting, overlooked in existing methods and settings.
    \item We propose a novel \dept\ framework, which introduces OOD detection into prompt tuning. Both our theoretical analysis and experimental results demonstrate the effectiveness of \dept\ framework for \setting.
    \item Based on \dept\ framework, we propose a novel prompt tuning approach \algo, which additionally enhances the base-class and new-class discriminability by introducing new-class detectors and sub-classifiers. 
    \item We conduct comprehensive experiments on \algo\ using 11 benchmark datasets. The results show that our proposed scheme outperforms current SOTA comparison methods, delivering a significant 2\% average improvement in accuracy.
\end{enumerate}

\section{Problem and Analysis}

In this section, we first describe the notions and problem formulation for the \setting\ setting. 
Subsequently, we conduct an empirical analysis using a real-world dataset~\cite{stanfordcars}, wherein we identify two primary challenges to address: base-to-new discriminability and new-class discriminability. 
Finally, we decompose the original problem to demonstrate that the incorporation of the OOD detection technique can effectively resolve these two challenges.

\subsection{Problem Formulation}

We focus on the prompt tuning setting for multi-class classification problems that involve an input space $\mathcal{X}$, a class space $\mathcal{Y} = \mathcal{Y}_{\bs} \cup \mathcal{Y}_{\nw} = [C]$, and the text space $\mathcal{T}$, where $C$ represents the number of classes. Here, $\mathcal{Y}_{\bs}$ denotes the set of base classes, and $\mathcal{Y}_{\nw}$ represents the set of new classes. The name of the $i$-th class is denoted as $\boldsymbol{t}_i \in \mathcal{T}$. Furthermore, $\boldsymbol{x} \in \mathcal{X}$ represents the data.
$f(\boldsymbol{x}) \in \mathcal{Y}$ and $g(\boldsymbol{x}) \in \{\bs, \nw\}$ denote the label of $\boldsymbol{x}$ and the specific class space to which it belongs, where $f$ and $g$ are the mapping functions of the ground truth of the labels and the class space.

In \setting\ problem, we are given a pre-trained vision-language model $\mathcal{F}=\{\text{E}_{V}, \text{E}_{T}\}$, which consists of a visual encoder $\text{E}_{V}: \mathcal{X} \mapsto \mathbb{R}^{d}$ and a textual encoder $\text{E}_{T}: \mathcal{T} \mapsto \mathbb{R}^{d}$, where $d$ represents the dimension of model $\mathcal{F}$. During the training stage, we learn the prompt vector $\boldsymbol{p}$ on a few-shot dataset $\mathcal{D}$ containing data derived from $\mathcal{Y}_{\bs}$. To simplify the notation, we define $t_i(\boldsymbol{p})$ as the concatenation of the tokens of the class name $\boldsymbol{t}_i$ and the learned prompt $\boldsymbol{p}$. Consequently, weight vectors $\{\boldsymbol{w}_i(\boldsymbol{p})\}_{i=1}^C$ are generated for each class as textual embeddings, where $\boldsymbol{w}_i(\boldsymbol{p}) = \text{E}_{T}\left (t_i(\boldsymbol{p}) \right ) / \| \text{E}_{T}\left (t_i(\boldsymbol{p}) \right ) \|$.
In the testing stage, given the test data $\boldsymbol{x}$ drawn from $\mathcal{Y}$, we initially obtain its visual embedding $\boldsymbol{z} = \text{E}_{V}(\boldsymbol{x}) / \| \text{E}_{V}(\boldsymbol{x}) \|$. Subsequently, we calculate the prediction probabilities as follows: 
\begin{equation}
P(y | \boldsymbol{x}) = \frac{ \exp{(\boldsymbol{z}^T \boldsymbol{w}_y / \tau)} }{ \sum_{i=1}^C \exp{(\boldsymbol{z}^T \boldsymbol{w}_i / \tau)} }
\end{equation} 
where $\tau$ represents the temperature determined by VLMs. For convenience, we will also use $P(\boldsymbol{x})$ to represent $P(y|\boldsymbol{x})$ in the subsequent paper. The prediction for $\boldsymbol{x}$ is given by $\mathop{\arg \max}\limits_{y \in \mathcal{Y}} P(y|\boldsymbol{x})$. The objective of \setting\ is to train a model that can make robust predictions on $\mathcal{Y}$, which includes both base and new classes, without experiencing overall performance degradation due to the presence of new classes. 
In our following analyses and experiments, we perform a comparison between the zero-shot baseline method (referred to as \zs) and the prompt tuning method (referred to as \pt) on \setting\ problem. 

\subsection{Problem Analysis}

\begin{figure}[t]
\begin{center}
	\subfloat[Zero-shot Baseline \zs]{\includegraphics[width=.45\columnwidth]{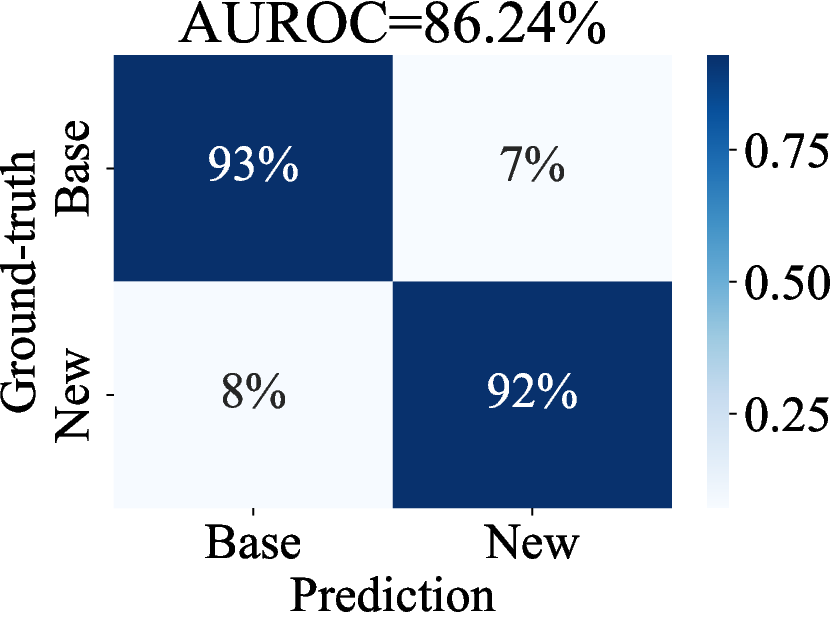}}\hspace{5pt}
	\subfloat[Prompt Tuning Method \pt]{\includegraphics[width=.45\columnwidth]{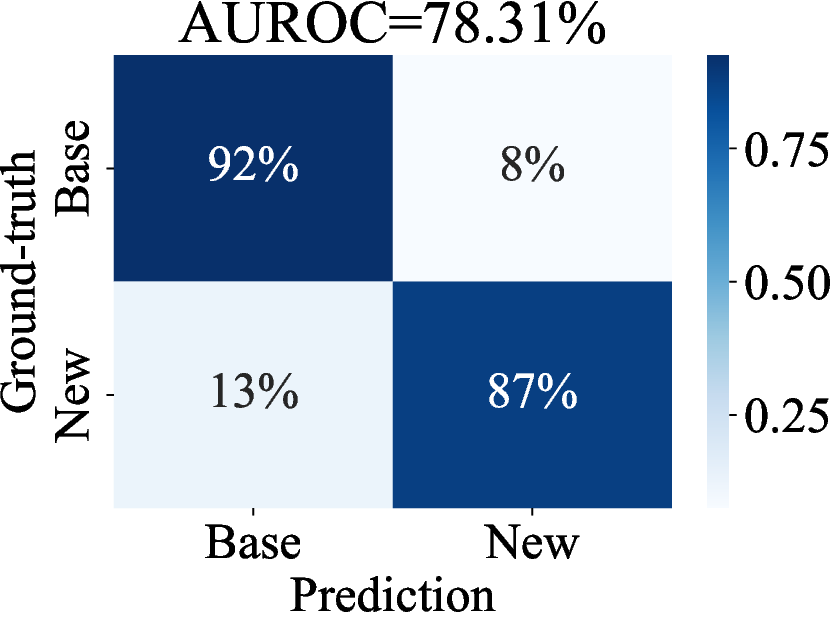}}
	\caption{Performance of \zs\ and \pt\ methods to distinguish data from base classes and new classes (base-to-new discriminability). }
    \label{fig: ProblemAnalysis-OOD}
\end{center}
\vskip -0.2in
\end{figure}

\begin{figure}[t]
    \begin{center}
        \subfloat[Zero-shot Baseline \zs]{\includegraphics[width=.45\columnwidth]{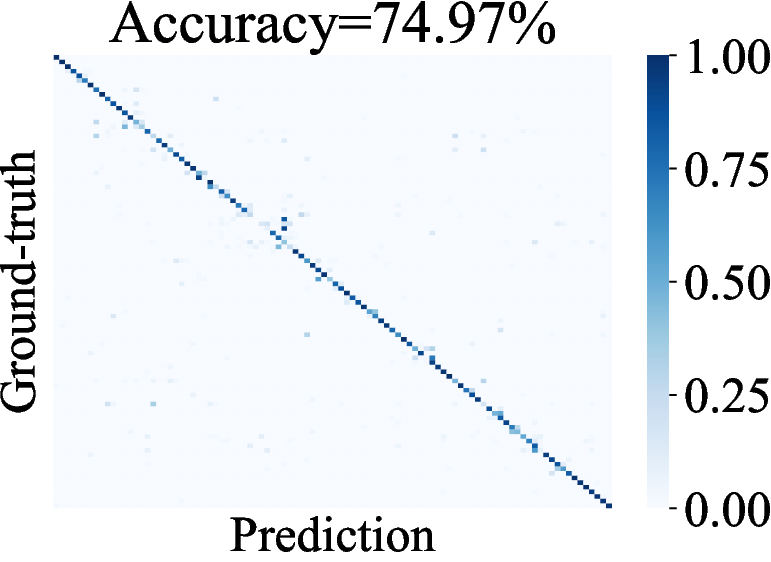}}\hspace{5pt}
        \subfloat[Prompt Tuning Method \pt]{\includegraphics[width=.45\columnwidth]{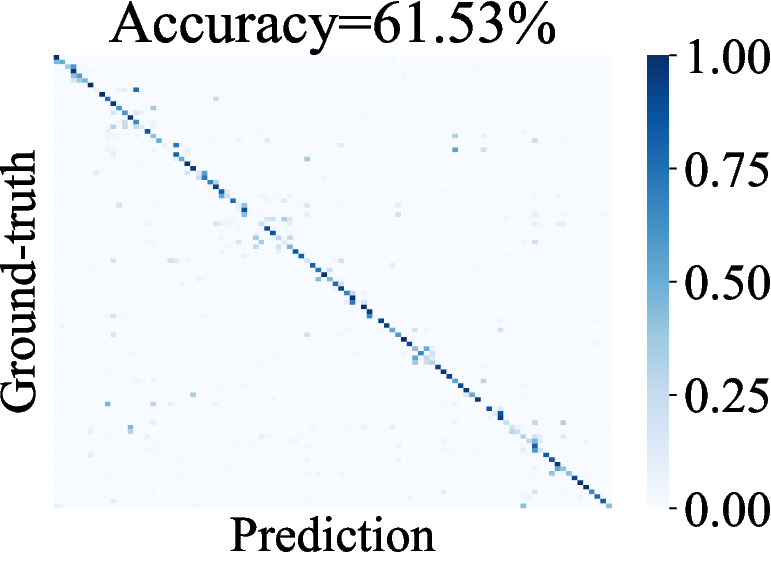}}
        \caption{Performance of \zs\ and \pt\ methods to distinguish data within new classes (new-class discriminability). }
        \label{fig: ProblemAnalysis-CLS}
    \end{center}
    \vskip -0.2in
\end{figure}

To tackle the \setting\ problem, we investigate a real-world dataset~\cite{stanfordcars} to conduct detailed analyses of the challenges inherent in \setting. 
Our observation demonstrates that while prompt tuning methods can improve 
base-class discriminability, they compromise both base-to-new discriminability and new-class discriminability. 
To illustrate this observation, we present a comparison between the \zs\ methods and \pt\ methods, where we employ \Clip\ as \zs\ method and \coop\ as \pt\ method, in Figures \ref{fig: ProblemAnalysis-OOD} and \ref{fig: ProblemAnalysis-CLS}.

\autoref{fig: ProblemAnalysis-OOD} indicates that the prompt tuning method results in a decreased base-to-new discriminability compared to the zero-shot baseline. Specifically, the AUROC for detecting new classes using the MSP technique~\cite{hendrycks2016baseline} decreases, and more false positive predictions are introduced for base classes. 
Moreover, \autoref{fig: ProblemAnalysis-CLS} illustrates that the prompt tuning method also exhibits reduced new-class discriminability compared to the zero-shot baseline.

We emphasize that the existing H metric is incapable of measuring base-to-new discriminability, making it unsuitable for comprehensive practical applications. In \setting\ problem, the accuracy evaluated in the entire class space can effectively address this limitation.

\subsection{Problem Decomposition}

The above analysis reveals that the zero-shot baseline surpasses the prompt tuning method in terms of both new-class discriminability and base-to-new discriminability. 
This observation motivates us to incorporate OOD detection technique to combine \zs\ method and \pt\ method. This approach aims to preserve the new-class discriminability using \zs\ while enhancing the base-class discriminability using \pt. 
Therefore, we decompose the original classification problem into separate OOD detection and two classification problems:
\begin{equation}
\begin{aligned}
    P \left ( y | \boldsymbol{x} \right ) 
    &= \sum\limits_{i \in \{\bs, \nw\}} P\left (y | y \in \mathcal{Y}_i, \boldsymbol{x} \right ) \cdot P\left (y \in \mathcal{Y}_i | \boldsymbol{x} \right ) \\
    &= P \left (y | y\in \mathcal{Y}_{k}, \boldsymbol{x} \right ) \cdot P\left (y \in \mathcal{Y}_{k} | \boldsymbol{x} \right ) 
    \label{eq: decomposition}
\end{aligned}
\end{equation}
where $k$ always equals $g(\boldsymbol{x})$ for the sake of simplicity, representing the ground-truth label space of $\boldsymbol{x}$. 
The second term is an OOD detector to determine whether $\boldsymbol{x}$ belongs to the base or new class space.
The first term is a classifier for the corresponding class space. 

\autoref{eq: decomposition} motivates us to propose a novel \emph{\textbf{De}composed \textbf{P}rompt \textbf{T}uning} framework (\dept), which synergistically leverages the advantages of both the zero-shot baseline \zs\ and the prompt tuning method \pt. The prediction probability $P_{\dept}(y|\boldsymbol{x})$ of \dept\ framework is:
\begin{equation}
    \begin{cases}
    P_{\pt}(y|\boldsymbol{x}), &P_{\ood}(y \in \mathcal{Y}_\bs |\boldsymbol{x}) \geq P_{\ood}(y \in \mathcal{Y}_\nw |\boldsymbol{x}), \\
    P_{\zs}(y|\boldsymbol{x}), &P_{\ood}(y \in \mathcal{Y}_\bs |\boldsymbol{x}) < P_{\ood}(y \in \mathcal{Y}_\nw |\boldsymbol{x}).
    \end{cases}
    \label{eq: dept}
\end{equation}
where $P_{\ood}(y \in \mathcal{Y}_b |\boldsymbol{x})$ is the OOD detector to determine whether $\boldsymbol{x}$ belongs to the base or new class space. 
$P_{\zs}(y|\boldsymbol{x})$ and $P_{\pt}(y|\boldsymbol{x})$ are classifiers of \zs\ and \pt. 
In following theoretical analysis and empirical experiment, we adopt the \zs\ method using MSP method as the OOD detector, i.e.,
$P_{\ood}(y \in \mathcal{Y}_i |\boldsymbol{x})= \max_{j \in \mathcal{Y}_i} P_{\zs}(y = j|\boldsymbol{x})$ for $i\in\{\bs, \nw\}$. 

Then, we adopt the cross-entropy metric of two probability distributions $\boldsymbol{p}$ and $\boldsymbol{q}$, i.e., $H(\boldsymbol{p}, \boldsymbol{q}) = -\sum_{i=1}^C p_i \log q_i$, to evaluate the performance of $P_{\zs}(y|\boldsymbol{x})$ and our \dept\ framework $P_{\dept}(y|\boldsymbol{x})$. 
We denote distributions $\tilde{\boldsymbol{k}} = \left \{\mathbb{I}[k = \bs], \mathbb{I}[k = \nw] \right \}$ 
and $\tilde{\boldsymbol{y}} = \left \{\mathbb[f(\boldsymbol{x})=i] \right \}_{i=1}^C$ for $\boldsymbol{x}$. 
Finally, we denote the following cross-entropy values for zero-shot baseline, prompt tuning method, and \dept\ framework: 
\begin{equation}
\begin{aligned}
    H^{\ood}_{\zs}(\boldsymbol{x}) 
    &= H \left (\tilde{\boldsymbol{k}}, \{P_{\zs}(y \in \mathcal{Y}_i| \boldsymbol{x})\}_{i=\{\bs, \nw\}} \right ), \\
    H^{\cls}_{\zs}(\boldsymbol{x}) 
    &= H \left (\tilde{\boldsymbol{y}}, \{P_{\zs}(y = j | y \in \mathcal{Y}_k, \boldsymbol{x})\}_{j=1}^C \right ), \\
    H^{\cls}_{\pt}(\boldsymbol{x}) 
    &= H \left (\tilde{\boldsymbol{y}}, \{P_{\pt}(y = j | y \in \mathcal{Y}_k, \boldsymbol{x})\}_{j=1}^C \right ), \\
    H_{\zs}(\boldsymbol{x})   
    &= H \left (\tilde{\boldsymbol{y}}, \{P_{\zs}(y = j| \boldsymbol{x})\}_{j=1}^C \right ), \\
    H_{\dept}(\boldsymbol{x}) 
    &= H \left (\tilde{\boldsymbol{y}}, \{P_{\dept}(y = j| \boldsymbol{x})\}_{j=1}^C \right ).
\end{aligned}
\end{equation}
\begin{theorem} 
\label{thm: error}
If $\mathbb{E}_{\boldsymbol{x}} \left [H^{\cls}_{\zs}(\boldsymbol{x}) \right ] \leq \delta$ for $\boldsymbol{x}$ belonging to both base and new classes, $\mathbb{E}_{\boldsymbol{x}} \left [H^{\cls}_{\pt}(\boldsymbol{x}) \right ] \leq \delta - \Delta$ for $\boldsymbol{x}$ belonging to base classes, and $\mathbb{E}_{\boldsymbol{x}} \left [H^{\ood}_{\zs}(\boldsymbol{x}) \right ] \leq \epsilon$, given a uniform mixing ratio ($\alpha: 1-\alpha$) of base classes and new classes in the testing data, we can determine that:
\begin{equation}
\begin{cases}
    \mathbb{E}_{\boldsymbol{x}} \left [ H_{\zs}(\boldsymbol{x}) \right ] &\leq \epsilon + \delta, \\
    \mathbb{E}_{\boldsymbol{x}} \left [ H_{\dept}(\boldsymbol{x}) \right ] &\leq \epsilon + \delta - \alpha \cdot \Delta .
\end{cases}    
\end{equation}
\end{theorem}
\begin{remark}
\autoref{thm: error} demonstrates that decomposing the zero-shot baseline into an OOD detector and classifiers, and incorporating prompt tuning methods to aid in classifying base classes, can effectively decrease the upper bound of classification error. Moreover, enhancing the reliability of the OOD detector helps reduce the error term $\epsilon$ and ensures that the performance on new classes remains uncompromised compared to the baseline method. 
Consequently, this framework preserves base-to-new discriminability and new-class discriminability of \zs\ method. 
Additionally, refining the \pt\ method increases $\Delta$, further enhancing base-class discriminability and reducing the upper bound of error.
\end{remark}
The proof is presented in \autoref{sec: proof_of_error}. \autoref{thm: error} motivates us to design a robust prompt tuning method based on \autoref{eq: dept} using OOD detection techniques to solve \setting. 
\section{\algo\ Approach}

We propose a novel prompt tuning framework, called \dept, to address the \setting\ problem. The \dept\ framework effectively maintains the discriminability between base classes and new classes, thus preventing degradation of discriminability when prompt tuning is applied. 
Our theoretical analysis, as presented in \autoref{thm: error}, demonstrates the superiority of \dept\ when combining the zero-shot baseline and prompt tuning method. 
However, there are still two challenges that need to be addressed in order to further enhance the performance in complex real-world applications: 
(1) How can we train reliable OOD detectors to identify new-class data using limited labeled data from base classes? 
(2) With reliable OOD detectors, how to separately improve the base-class and new-class discriminability?

\begin{figure*}[t]
    \begin{center}
    \centerline{\includegraphics[width=\linewidth]{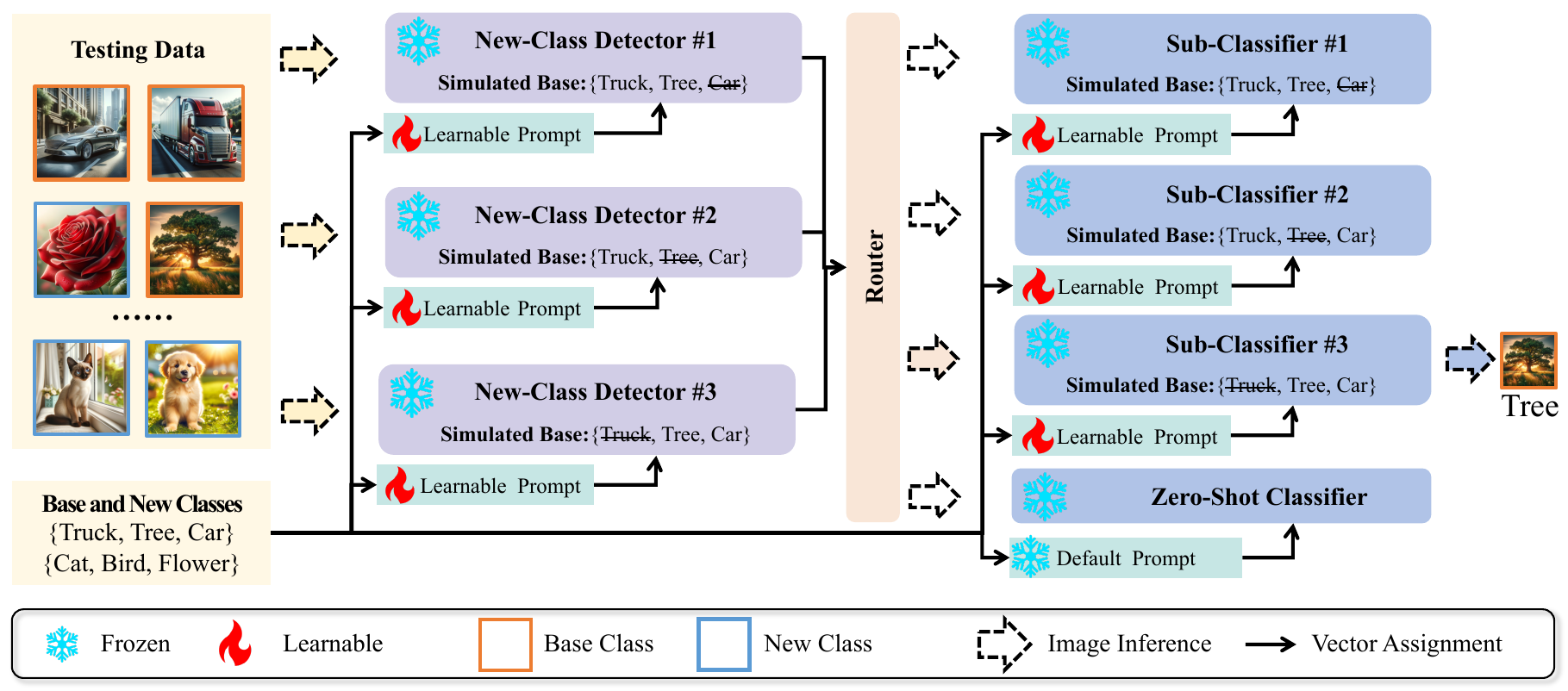}}
    \caption{The overall illustration of \algo\ approach.}
    \label{fig: decoop}
    \end{center}
    \vskip -0.2in
\end{figure*}

To tackle the challenges above, we present a novel prompt tuning approach named \emph{\textbf{De}composed \textbf{Co}ntext \textbf{Op}timization} (\algo) based on our \dept\ framework, containing $K$ new-classes detectors $\{\mathcal{M}_D^i\}_{i=1}^K$ and sub-classifiers $\{\mathcal{M}_C^i\}_{i=1}^K$. 
The introduction of new-class detectors aids in the improved detection of data from new classes in \setting\ problem, where the names of new classes are known and can be utilized. This differs from the traditional OOD detection problems and presents an opportunity for further performance enhancement.
The sub-classifiers are designed to better classify the data from base classes and reduce the potential risks for new classes, which aims to enhance the base-class and new-class discriminability with a reliable base-to-new discriminability. 
The overall illustration of \algo\ approach is shown in \autoref{fig: decoop} and each component is described thoroughly in the following subsections.

\subsection{New-class Detector $\mathcal{M}_D$}

In the \setting\ problem, the model is trained with $\mathcal{Y}_\bs$ but has knowledge of the entire class space $\mathcal{Y}$ during testing. Therefore, the main challenge for new class detectors is to train the model to effectively utilize the knowledge of the new class $\mathcal{Y}_\nw$, which is only known during testing.

Specifically, Our proposed solution incorporates a leave-out strategy which divides the base class space $\mathcal{Y}_\bs$ into two distinct subsets during training stage: simulated base classes $\hat{\mathcal{Y}}_\bs$ and simulated new classes $\hat{\mathcal{Y}}_\nw$, where $\hat{\mathcal{Y}}_\bs \cup \hat{\mathcal{Y}}_\nw = \mathcal{Y}_\bs$. Respectively, we split the original training set $\mathcal{D}$ into $\mathcal{D}_\bs = \{(\boldsymbol{x}, y)|(\boldsymbol{x}, y)\sim \mathcal{D} \land y \in \hat{\mathcal{Y}}_\bs \}$ and $\mathcal{D}_\nw = \{(\boldsymbol{x}, y) |(\boldsymbol{x}, y)\sim\mathcal{D} \land y \in \hat{\mathcal{Y}}_\nw \}$. 
Then, our optimization objective function for the new-class detector is defined as:
\begin{equation}
\begin{aligned}
\ell_{\ood}
= & \frac{1}{|\mathcal{D}_\bs|} \sum\limits_{(\boldsymbol{x}, y) \sim \mathcal{D}_\bs} \ell_{CE}(\boldsymbol{x}, y) \\
  & + \max \left \{0, \gamma + \ell_{E}^\bs - \ell_{E}^\nw \right \}
\end{aligned}
\label{eq: ood-loss}
\end{equation}
where $\ell_{CE}(\boldsymbol{x}, y) = -\log{P(\boldsymbol{x})_y}$ represents the cross-entropy loss, $\ell_{E}(\boldsymbol{x}) = -\sum_{i=1}^C P(\boldsymbol{x})_i \log{P(\boldsymbol{x})_i}$ represents the entropy loss, $\ell_{E}^\bs = \frac{1}{|\mathcal{D}_\bs|} \sum_{(\boldsymbol{x}, y) \sim \mathcal{D}_\bs} \ell_{E}(\boldsymbol{x})$ represents the average entropy on the simulated base classes, and $\ell_{E}^\nw = \frac{1}{|\mathcal{D}_\nw|} \sum_{(\boldsymbol{x}, y) \sim \mathcal{D}_\nw} \ell_{E}(\boldsymbol{x})$ represents the average entropy on the simulated new classes. Additionally, $\gamma$ is a hyperparameter that controls the margin between $\ell_{E}^\bs$ and $\ell_{E}^\nw$ to ensure stable optimization. The objective function in \autoref{eq: ood-loss} encourages the model to make low-entropy predictions on simulated base classes and high-entropy predictions on simulated new classes, thereby enhancing base-to-new discriminability. 
However, partitioning the base class space causes the model's cognition to be limited to a subset of base classes, leading to the failure to distinguish between other base classes and new classes during testing. To address this issue, we propose the adoption of an ensemble of $K$ new-class detectors $\{\mathcal{M}_D^i\}_{i=1}^K$ that cover the entire base class space during training.
Each new-class detector is trained with \autoref{eq: ood-loss} with a different class partition.
Our class partitions of $K$ new-class detectors ensure each base class is considered as a simulated new class for at least one new-class detector. We denote $\mathcal{M}_D^i(\boldsymbol{x})$ as the new-class score computed for $\boldsymbol{x}$. Lower scores indicate a higher likelihood that $\boldsymbol{x}$ belongs to new classes.

In addition, a threshold remains crucial for the detection of new classes, even when well-trained new-class detectors are provided. Leveraging the benefits of our partition and ensemble strategy, we can directly estimate the threshold for each new-class detector during training using the Otsu algorithm~\cite{Otsu79, LiuY09} and training data.
This is possible due to the presence of naturally simulated base classes and new classes in the training data for each new-class detector. 
Subsequently, these estimated thresholds can be averaged to determine the threshold value, denoted as $\tau$, for testing.

\subsection{Sub-Classifier $\mathcal{M}_C$}

After training reliable new-class detectors, we proceed to train a sub-classifier for each detector, as each detector focuses on a specific subset of the base class space. 
Each of the $K$ sub-classifiers, denoted as $\{\mathcal{M}_C^i\}_{i=1}^K$, is designed to specialize in a particular base class space, thereby achieving better discriminability for the corresponding subset class space. 
For the $i$-th sub-classifier $\mathcal{M}_C^i$, we first utilize the trained new-class detector $\mathcal{M}_D^i$  partition the training data into two subsets: $\mathcal{D}_\bs^i$ and $\mathcal{D}_\nw^i$. 
Here, $\mathcal{D}_\bs^i = \left \{(\boldsymbol{x}, y)|(\boldsymbol{x}, y)\sim\mathcal{D} \land \mathcal{M}_D^i(\boldsymbol{x}) \geq \tau \right \}$ and $\mathcal{D}_\nw^i = \left \{(\boldsymbol{x}, y) |(\boldsymbol{x}, y)\sim\mathcal{D} \land \mathcal{M}_D^i(\boldsymbol{x}) < \tau \right \}$.
Subsequently, we optimize the following objective function:
\begin{equation}
\ell_{\cls} = \sum\limits_{(\boldsymbol{x}, y)\sim \mathcal{D}_\bs^i}\ell_{CE}(\boldsymbol{x}, y) 
+ \sum\limits_{(\boldsymbol{x}, y)\sim \mathcal{D}_\nw^i}\ell_{KL}\left (P(\boldsymbol{x}), P_{\zs}(\boldsymbol{x}) \right )
\label{eq: iid-loss}
\end{equation}
Here, $\ell_{KL}$ denotes KL-divergence loss, and $P(\boldsymbol{x})$ and $P_{\zs}(\boldsymbol{x})$ represent the prediction probabilities of \algo\ approach and zero-shot CLIP baseline. 
We denote $\mathcal{M}_C^i(\boldsymbol{x})$ as the prediciton probabilities computed for $\boldsymbol{x}$.

\subsection{Inference}

During testing, we evaluate an ensemble of $K$ new-class detectors $\{\mathcal{M}_D^i\}_{i=1}^K$ to determine whether each testing data should be predicted by one of the learned sub-classifiers $\mathcal{M}_C^i$ or the zero-shot CLIP baseline. Specifically, for a testing instance $\boldsymbol{x}$, we first compute the scores of the new-class detectors, $\{\mathcal{M}_D^i(\boldsymbol{x})\}_{i=1}^K$, and then make the prediction according to our \algo\ approach, defined as:
\begin{equation}
P_{\algo}(\boldsymbol{x}) =
\begin{cases}
P_{\zs}(\boldsymbol{x}), &\text{if } \mathop{\max}\limits_{i \in \{1, \cdots, K\}} \mathcal{M}_D^i(\boldsymbol{x}) < \tau, \\
\mathcal{M}_C^{i^{\star}}(\boldsymbol{x}), &\text{if } \mathop{\max}\limits_{i \in \{1, \cdots, K\}} \mathcal{M}_D^i(\boldsymbol{x}) \geq \tau,
\end{cases}
\end{equation}
where $i^{\star} = \mathop{\arg \max}_{i \in \{1, \cdots, K\}} \mathcal{M}_D^i(\boldsymbol{x})$. 
\algo\ approach selects single sub-classifier to predict each testing data instead of aggregating the results from all sub-classifiers. As a result, our approach requires $K$ times computation for the new-class detectors compared to the zero-shot CLIP baseline. In our experiments, we set $K$ to 3, which does not impose a heavy computational burden. We conduct experiments about evaluation time in Appendix \ref{sec:time-exps}, demonstating that \algo\ is relatively efficient. 
\section{Experiments}

In this section, we conduct experiments to answer the following three research questions:
\begin{enumerate}[] 
    \item \textbf{RQ1}: Can the empirical results of the \dept\ framework on real-world datasets conform to our theoretical analysis? 
    \item \textbf{RQ2}: Can the \algo\ method surpass existing baseline and SOTA methods, thereby demonstrating its robustness? 
    \item \textbf{RQ3}: Does the \algo\ successfully improve the base-to-new discriminability, as designed? 
\end{enumerate}

\begin{table}[t]
    \centering
    \caption{Comparison of average performance across 11 datasets was conducted among three approaches: \zs, \pt, and our \dept\ framework, utilizing ViT-B/16 and ViT-B/32 architectures. These results are consistent with our theoretical analysis.}
    \label{tab:theorem-exps}
    \vskip 0.1in
    \begin{center}
    \begin{small}
    \begin{sc}
    \resizebox{\linewidth}{!}{
    \begin{tabular}{l|cc|cc}
    \toprule
    \midrule
    \multirow{2}{*}{Method} & \multicolumn{2}{c|}{ViT-B/16} & \multicolumn{2}{c}{ViT-B/32}\\ \cmidrule{2-5} 
                            & New Acc.    & Accuracy  & New Acc. & Accuracy   \\ \midrule
    \zs         & 65.49 & 63.92 & 63.95 & 60.36 \\
    \pt         & 57.73 & 65.57 & 53.01 & 61.03 \\
    \dept       & \textbf{68.15} & \textbf{68.03} & \textbf{65.45} & \textbf{62.92}\\
    \midrule
    \bottomrule  
    \end{tabular}}
    \end{sc}
    \end{small}
    \end{center}
    \vskip -0.1in
\end{table}

\begin{table}[t]
    \centering
    \caption{The average performance across 11 datasets using ViT-B/16 and ViT-B/32 architectures. The best performance is in bold. }
    \label{tab:average-exps}
    \vskip 0.1in
    \begin{center}
    \begin{small}
    \begin{sc}
    \resizebox{\linewidth}{!}{
    \begin{tabular}{l|cc|cc}
    \toprule
    \midrule
    \multirow{2}{*}{Method} & \multicolumn{2}{c|}{ViT-B/16} & \multicolumn{2}{c}{ViT-B/32}\\ \cmidrule{2-5} 
          & H     &Accuracy& H     & Accuracy   \\ \midrule
    \Clip & 70.84 & 63.92  & 67.13 & 60.36  \\ 
    \Pens & 71.65 & 65.39  & 67.76 & 60.73  \\ 
    \coop & 72.14 & 65.57  & 67.86 & 61.03  \\ 
    \cocoop & 74.72 & 67.67  & 70.77 & 62.96  \\ 
    \ship & 72.26 & 64.51  & 69.25 & 59.91  \\ 
    \algo(Ours) & \textbf{76.13} & \textbf{69.69}  & \textbf{72.51} & \textbf{65.75}  \\ 
    \midrule
    \bottomrule  
    \end{tabular}}
    \end{sc}
    \end{small}
    \end{center}
    \vskip -0.1in
\end{table}

\begin{table*}[t]
    \centering
    \caption{Performance comparison on 11 datasets using ViT-B/16 architecture. The best performance is in bold.}
    \label{tab:full-exps-vit16}
    \vskip 0.1in
    \begin{center}
    \begin{small}
    \begin{sc}
    \resizebox{\textwidth}{!}{
    \begin{tabular}{l|cc|ll|ll|ll}
    \toprule
    \midrule
                    & \multicolumn{2}{c|}{Average}                            & \multicolumn{2}{c|}{ImageNet}                           & \multicolumn{2}{c|}{Caltech101}                         & \multicolumn{2}{c}{OxfordPets}                          \\
                    & \multicolumn{1}{c}{H}      & \multicolumn{1}{c|}{Acc.}  & \multicolumn{1}{c}{H}      & \multicolumn{1}{c|}{Acc.}  & \multicolumn{1}{c}{H}      & \multicolumn{1}{c|}{Acc.}  & \multicolumn{1}{c}{H}      & \multicolumn{1}{c}{Acc.}   \\ \midrule 
    \Clip           & 70.84                      & 63.92                      & 70.20 $\pm$ 0.00           & 66.73 $\pm$ 0.00           & 95.41 $\pm$ 0.00           & 92.90 $\pm$ 0.00           & 92.93 $\pm$ 0.00           & 88.03 $\pm$ 0.00           \\ 
    \Pens           & 71.65                      & 65.39                      & 72.00 $\pm$ 0.00           & 68.48 $\pm$ 0.00           & 96.20 $\pm$ 0.00           & 94.08 $\pm$ 0.00           & 92.42 $\pm$ 0.00           & 86.37 $\pm$ 0.00           \\ 
    \coop           & 72.14                      & 65.57                      & 64.95 $\pm$ 1.11           & 61.79 $\pm$ 1.09           & 95.96 $\pm$ 0.39           & 93.24 $\pm$ 0.68           & 95.38 $\pm$ 0.33           & 89.61 $\pm$ 0.34           \\ 
    \cocoop         & 74.72                      & 67.67                      & 72.71 $\pm$ 0.33           & 69.41 $\pm$ 0.36           & 95.55 $\pm$ 0.24           & 93.43 $\pm$ 0.37           & \textbf{95.71 $\pm$ 0.76}  & \textbf{90.24 $\pm$ 1.32}  \\ 
    \ship           & 72.26                      & 64.51                      & 67.29 $\pm$ 0.38           & 63.65 $\pm$ 0.32           & 95.83 $\pm$ 0.23           & 92.93 $\pm$ 0.37           & 94.44 $\pm$ 0.54           & 86.78 $\pm$ 1.32           \\ 
    \algo(Ours)     & \textbf{76.13 }            & \textbf{69.69}             & \textbf{72.98 $\pm$ 0.04}  & \textbf{69.62 $\pm$ 0.08}  & \textbf{96.52 $\pm$ 0.09}  & \textbf{94.50 $\pm$ 0.22}  & 95.27 $\pm$ 0.08           & 88.87 $\pm$ 0.28           \\ 
    \midrule
    \midrule
                    & \multicolumn{2}{c|}{StandfordCars}                      & \multicolumn{2}{c|}{Flowers102}                         & \multicolumn{2}{c|}{Food101}                            & \multicolumn{2}{c}{FGVCAircraft}                        \\
                    & \multicolumn{1}{c}{H}      & \multicolumn{1}{c|}{Acc.}  & \multicolumn{1}{c}{H}      & \multicolumn{1}{c|}{Acc.}  & \multicolumn{1}{c}{H}      & \multicolumn{1}{c|}{Acc.}  & \multicolumn{1}{c}{H}      & \multicolumn{1}{c}{Acc.}   \\ \midrule 
    \Clip           & 68.75 $\pm$ 0.00           & 65.39 $\pm$ 0.00           & 72.74 $\pm$ 0.00           & 67.28 $\pm$ 0.00           & 90.18 $\pm$ 0.00           & 85.40 $\pm$ 0.00           & 30.25 $\pm$ 0.00           & 23.94 $\pm$ 0.00           \\ 
    \Pens           & 69.36 $\pm$ 0.00           & 65.95 $\pm$ 0.00           & 72.14 $\pm$ 0.00           & 67.03 $\pm$ 0.00           & 90.32 $\pm$ 0.00           & 85.54 $\pm$ 0.00           & 29.42 $\pm$ 0.00           & 23.31 $\pm$ 0.00           \\ 
    \coop           & 68.22 $\pm$ 0.49           & 63.81 $\pm$ 0.44           & 78.33 $\pm$ 2.26           & 72.11 $\pm$ 2.36           & 86.65 $\pm$ 1.38           & 80.84 $\pm$ 1.50           & 29.38 $\pm$ 1.78           & 24.80 $\pm$ 1.23           \\ 
    \cocoop         & 71.49 $\pm$ 0.62           & 67.75 $\pm$ 0.68           & 80.04 $\pm$ 1.46           & 71.95 $\pm$ 1.24           & 90.41 $\pm$ 0.24           & 85.61 $\pm$ 0.43           & 27.87 $\pm$ 11.36          & 21.46 $\pm$ 7.42           \\ 
    \ship           & 69.71 $\pm$ 0.43           & 64.67 $\pm$ 0.55           & 76.85 $\pm$ 2.18           & 70.40 $\pm$ 2.01           & 86.84 $\pm$ 1.49           & 77.39 $\pm$ 2.19           & 27.13 $\pm$ 1.10           & 24.44 $\pm$ 0.96           \\ 
    \algo(Ours)     & \textbf{73.24 $\pm$ 0.15}  & \textbf{69.64 $\pm$ 0.19}  & \textbf{84.16 $\pm$ 0.27}  & \textbf{78.61 $\pm$ 0.59 } & \textbf{90.68 $\pm$ 0.09}  & \textbf{85.83 $\pm$ 0.07}  & \textbf{31.44 $\pm$ 0.39}  & \textbf{25.15 $\pm$ 0.31}  \\ 
    \midrule
    \midrule
                    & \multicolumn{2}{c|}{SUN397}                             & \multicolumn{2}{c|}{DTD}                                & \multicolumn{2}{c|}{EuroSAT}                            & \multicolumn{2}{c}{UCF101}                              \\
                    & \multicolumn{1}{c}{H}      & \multicolumn{1}{c|}{Acc.}  & \multicolumn{1}{c}{H}      & \multicolumn{1}{c|}{Acc.}  & \multicolumn{1}{c}{H}      & \multicolumn{1}{c|}{Acc.}  & \multicolumn{1}{c}{H}      & \multicolumn{1}{c}{Acc.}   \\ \midrule 
    \Clip           & 72.26 $\pm$ 0.00           & 62.57 $\pm$ 0.00           & 57.32 $\pm$ 0.00           & 44.56 $\pm$ 0.00           & 58.16 $\pm$ 0.00           & 41.40 $\pm$ 0.00           & 71.00 $\pm$ 0.00           & 64.97 $\pm$ 0.00           \\ 
    \Pens           & 75.04 $\pm$ 0.00           & 65.97 $\pm$ 0.00           & 59.63 $\pm$ 0.00           & 46.28 $\pm$ 0.00           & 58.45 $\pm$ 0.00           & 48.91 $\pm$ 0.00           & 73.17 $\pm$ 0.00           & 67.33 $\pm$ 0.00           \\ 
    \coop           & 71.37 $\pm$ 1.21           & 61.82 $\pm$ 1.11           & 57.22 $\pm$ 2.37           & 48.18 $\pm$ 1.78           & 74.33 $\pm$ 4.35           & 59.65 $\pm$ 5.07           & 71.68 $\pm$ 2.84           & 65.41 $\pm$ 2.18           \\ 
    \cocoop         & 77.17 $\pm$ 0.27           & 68.17 $\pm$ 0.33           & 60.59 $\pm$ 1.51           & 47.90 $\pm$ 1.43           & 73.77 $\pm$ 3.58           & 58.08 $\pm$ 1.49           & 76.59 $\pm$ 0.79           & 70.39 $\pm$ 1.25           \\ 
    \ship           & 72.57 $\pm$ 0.38           & 60.42 $\pm$ 0.48           & 56.82 $\pm$ 2.18           & 47.58 $\pm$ 1.62           & 73.29 $\pm$ 2.67           & 54.11 $\pm$ 1.73           & 74.09 $\pm$ 2.09           & 67.24 $\pm$ 1.94           \\ 
    \algo(Ours)     & \textbf{78.11 $\pm$ 0.09}  & \textbf{69.33 $\pm$ 0.05}  & \textbf{62.72 $\pm$ 1.23}  & \textbf{51.44 $\pm$ 1.04}  & \textbf{74.61 $\pm$ 3.82}  & \textbf{61.90 $\pm$ 3.72}  & \textbf{77.67 $\pm$ 0.50}  & \textbf{71.71 $\pm$ 0.79}  \\ 
    \midrule
    \bottomrule
    \end{tabular}
    }
    \end{sc}
    \end{small}
    \end{center}
    \vskip -0.1in
\end{table*}

\subsection{Experimental Setup}

\paragraph{Evaluation Protocol.} 
We adopt the few-shot prompt tuning setting as previously explored in studies such as \cite{clip, cocoop, ship}. This setting involves partitioning the class space of each dataset equally, with 50\% of the classes designated as base classes and the remaining 50\% as new classes. 
Consequently, for each dataset, prompts are learned for downstream tasks using 16 labeled samples per base class, drawn from the training set. The efficacy of these learned prompts is subsequently evaluated on the entire testing set, encompassing both base and new classes.
In \algo\ method, we report the Accuracy as well as previously reported H metric. 
As per the definition in CoCoOp~\cite{cocoop}, H metric separately evaluates the accuracy on base classes and new classes, denoted as $\text{Acc}_{\text{base}}$ and $\text{Acc}_{\text{new}}$. Then, H metric is computed using their harmonic mean, defined as $\text{H} = \frac{2 \times \text{Acc}_{\text{base}} \times \text{Acc}_{\text{new}}}{\text{Acc}_{\text{base}} + \text{Acc}_{\text{new}}}$. 
The metric H evaluates the overall performance of classifying both base and new classes separately, which we refer to as base-class discriminability and new-class discriminability. 
We evaluate the accuracy of the entire class space, which includes a mix of base and new classes, denoted as Accuracy. 
This metric evaluates the overall performance of classifying both base and new classes, while additionally measuring base-to-new discriminability compared to the H metric.

\paragraph{Datasets.} 
Following the CoOp framework~\cite{coop}, we conducted evaluations of our proposed \algo\ framework along with comparison methods on various image classification tasks. 
These tasks included general object recognition using ImageNet~\cite{imagenet} and Caltech-101~\cite{caltech101} datasets, fine-grained object recognition involving datasets such as Oxford Pets~\cite{stanfordcars}, Food-101~\cite{food101}, Stanford Cars~\cite{stanfordcars}, Oxford Flowers 102~\cite{oxfordflowers}, and FGVC Aircraft~\cite{fgvc}. Additionally, we performed a remote sensing recognition task using the EuroSAT~\cite{eurosat} dataset, a texture recognition task using the DTD~\cite{dtd} dataset, an action recognition task using UCF101~\cite{ucf101} dataset and a large-scale scene understanding task using SUN397~\cite{sun397} dataset.
For each dataset, we developed a few-shot training set for prompt tuning and employed the full testing set to evaluate the effectiveness of the learned prompts.

\paragraph{Compared Methods.} 
We compare our approach with five existing prompt-based methods. 
\Clip~\cite{clip} uses a hand-crafted prompt to generate the target classifier on the downstream task. 
Furthermore, we compare the \Pens\ method, an ensemble technique that utilizes multiple classifiers to enhance the performance of \Clip, adhering to the guidelines set by CLIP. 
\coop~\cite{coop} learns a soft prompt by minimizing the classification loss, and \cocoop~\cite{coop} extends \coop  \ by further learning a lightweight neural network to generate for each image an input-conditional token. 
\ship~\cite{ship} follows variational autoencoders to introduce a generator that reconstructs the visual features by inputting the synthesized prompts and the corresponding class names to the textual encoder of CLIP. 

\paragraph{Implementation Details.}

The number of tokens in each prompt is set to 16 for \algo\ approach and comparison methods. 
We train the prompts of new-class detectors for 50 epochs using the SGD optimizer and subsequently train the prompts for sub-classifiers for 100 epochs, also using the SGD optimizer. The learning rate $lr$ is set to 0.002, and it follows a cosine decay schedule. The margin $\gamma$ is set to $0.4$ for all datasets. We use the \Pens\ method as our zero-shot baseline within the \algo\ approach.
The batch size for images is 32 across all datasets. All experiments were conducted on Linux servers equipped with NVIDIA A800 GPUs. We report the average results over 5 runs with different random seed $\{1,2,3,4,5\}$.

\subsection{Empirical Results}

\textbf{RQ1}: Can the empirical results of the \dept\ framework on real-world datasets conform to our theoretical analysis? 

To verify \autoref{thm: error}, we conducted experiments on 11 datasets using ViT-B/16 and ViT-B/32 architectures. 
We employed \Clip\ as the zero-shot baseline \zs \ and \coop\ as the prompt tuning method \pt. 
Subsequently, we constructed our \dept\ framework by integrating these two methods, as presented in \autoref{eq: dept}. 
Here, the OOD detector used in our \dept\ framework directly derives from \Clip\ using MSP method~\cite{hendrycks2016baseline}. 
Each method is evaluated on the entire class space  $\mathcal{Y}$, and the average performance across all datasets is reported. The results include New Acc. and Accuracy, indicating the average performance of new classes and all classes, respectively.
The results presented in \autoref{tab:theorem-exps} consistently demonstrate that our \dept\ framework outperforms both \zs\ and \pt\ methods when evaluated using the New Acc. and Accuracy metrics. This observation suggests that the \dept\ framework effectively mitigates performance degradation on new classes through the utilization of the OOD detector, which aligns well with our theoretical analysis.

\textbf{RQ2}: Can the \algo\ method surpass existing baseline and SOTA methods, thereby demonstrating its robustness? 

To assess the effectiveness of the \algo\ approach, we conducted experiments on 11 datasets using ViT-B/16 and ViT-B/32 architectures. The average performance across all datasets, as well as the detailed performance on each dataset measured by two metrics, i.e., H and Accuracy, is reported. 
The results obtained using the ViT-B/16 architecture are presented in \autoref{tab:full-exps-vit16}. 
Our \algo\ approach demonstrates superior average performance on both the H metric and Accuracy, showcasing its robustness. 
Regarding the detailed performance on each dataset, our approach outperforms the comparison methods on 10 out of 11 datasets, while achieving comparable performance on the remaining dataset. 
The detailed results using the ViT-B/32 architecture are provided in Appendix \ref{sec:full-exps-vit32}, which yield similar conclusions.

Furthermore, these results reveal a positive correlation between the H metric and Accuracy in most cases. However, specific datasets such as FGVCAircraft~\cite{fgvc} show that higher H metric values do not necessarily lead to improved Accuracy. This observation suggests that the H metric is inadequate for measuring base-to-new discriminability, emphasizing the significance of \setting\ problem.

\textbf{RQ3}: Does the \algo\ successfully improve the base-to-new discriminability, as designed? 

\begin{table}[t]
    \centering
    \caption{The base-to-new discriminability of each method evaluated using MSP method~\cite{hendrycks2016baseline} and AUROC metrics. The best performance is in bold.}
    \label{tab:auroc-exps}
    \vskip 0.1in
    \begin{center}
    \begin{small}
    \begin{sc}
    \resizebox{\linewidth}{!}{
    \begin{tabular}{l|cccc}
    \toprule
    \midrule
    dataset & \Clip & \cocoop & \ship & \algo(Ours)  \\
    \midrule
    ImageNet        & 88.34 & 88.05 & 84.71 & \textbf{97.48} \\ 
    Caltech101      & 97.03 & 95.71 & 96.94 & \textbf{99.58} \\ 
    OxfordPets      & 92.66 & 91.15 & 93.30 & \textbf{98.12} \\ 
    StanfordCars    & 86.24 & 83.00 & 87.23 & \textbf{97.63} \\ 
    Flowers102      & 84.92 & 79.63 & 84.84 & \textbf{95.75} \\ 
    Food101         & 89.88 & 88.19 & 89.92 & \textbf{97.59} \\ 
    FGVCAircraft    & 75.08 & 69.00 & 75.78 & \textbf{84.06} \\ 
    SUN397          & 72.46 & 73.75 & 74.78 & \textbf{90.21} \\ 
    DTD             & 62.29 & 60.65 & 60.66 & \textbf{75.47} \\ 
    EuroSAT         & 56.40 & 57.74 & 59.32 & \textbf{77.78} \\ 
    UCF101          & 82.03 & 79.03 & 80.35 & \textbf{93.56} \\ \hline
    Average         & 80.67 & 78.72 & 80.71 & \textbf{91.57} \\ 
    \midrule
    \bottomrule  
    \end{tabular}}
    \end{sc}
    \end{small}
    \end{center}
    \vskip -0.1in
\end{table}

\begin{figure}[t]
    \begin{center}
    \centerline{\includegraphics[width=\columnwidth]{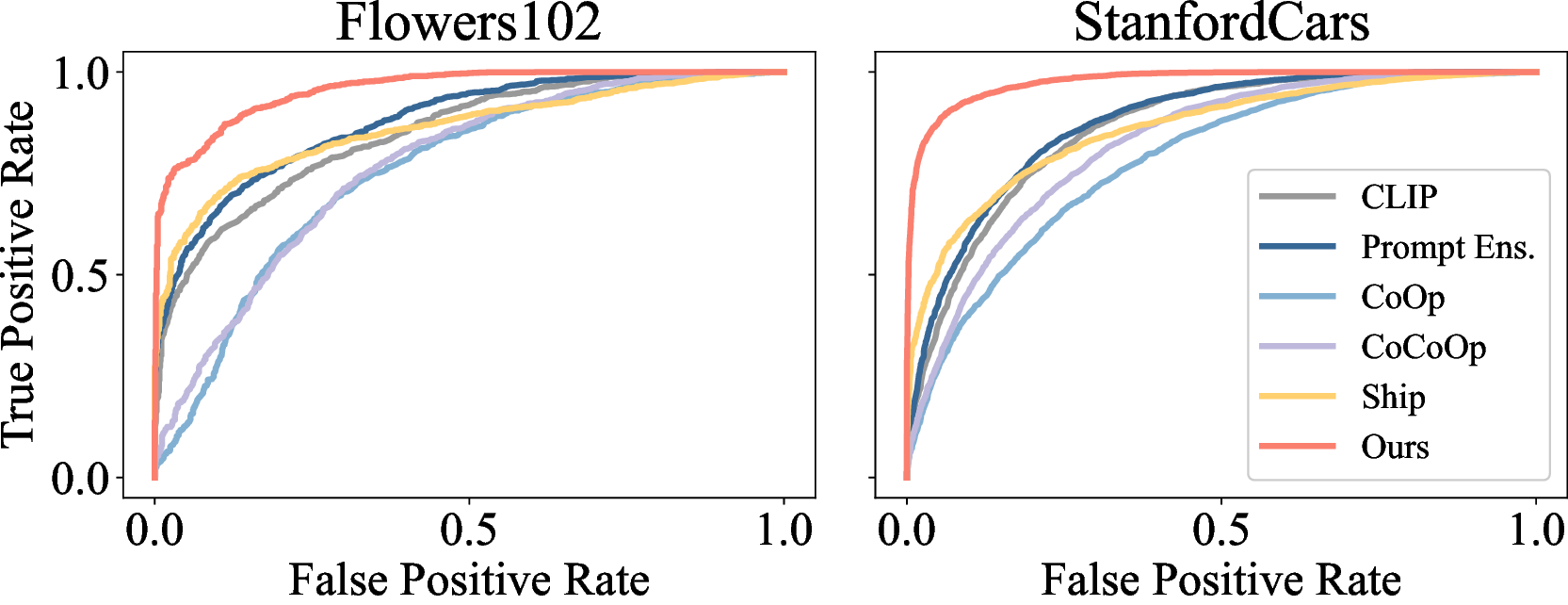}}
    \caption{The ROC curve for detecting new classes of each method on Flowers102 and StandfordCars datasets.}
    \label{fig: select-auroc}
    \end{center}
    \vskip -0.2in
\end{figure}

The \algo\ approach introduces new-class detectors with the aim of improving base-to-new discriminability while simultaneously enhancing the discriminability of new classes. 
We evaluate the base-to-new discriminability of our approach and selected methods using the MSP~\cite{hendrycks2016baseline} method with the ViT-B/16 architecture. 
Specifically, for each method, we use the maximum probability on base classes as the scores and report the AUROC~\cite{Bradley97} in \autoref{tab:auroc-exps}. The results clearly indicate that our \algo\ approach significantly improves base-to-new discriminability, which accounts for its SOTA performance. We have omitted some methods and standard deviations due to space limitations. Please refer to Appendix \ref{sec:full-exps-auroc} for full results. 
Additionally, we present the ROC curves for two representative datasets in \autoref{fig: select-auroc}, which demonstrates similar findings. Due to space limitations, the ROC curves for all datasets are provided in Appendix \ref{sec:full-exps-roc}. Furthermore, we explore the correlation between the performance of new-class detectors and the model in Appendix \ref{sec:correlation}.

\begin{figure}[t]
\begin{center}
\includegraphics[width=\linewidth]{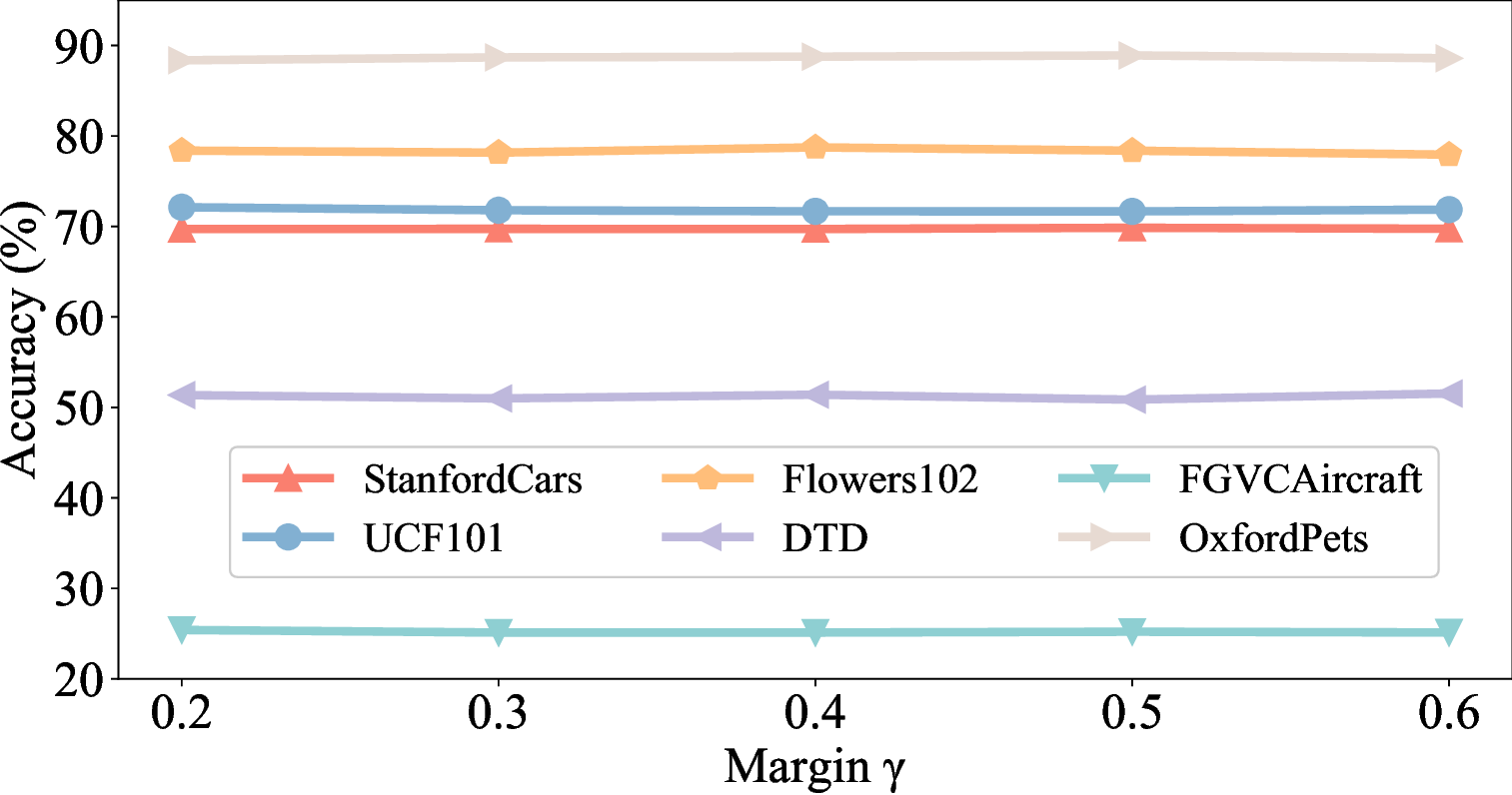}
\caption{Performance using different values of margin $\gamma$.}
\label{fig: margin}
\end{center}
\vskip -0.2in
\end{figure}

\paragraph{Hyperparemeter. } 
The margin $\gamma$ serves as a hyperparameter for learning new-class detectors in our \algo\ approach. It controls the margin in the optimization process of the detectors, which may affect their performance.
To answer the robustness question of $\gamma$, we conduct experiments on six datasets. \autoref{fig: margin} demonstrates the robustness of the \algo\ approach to changes in $\gamma$.

\paragraph{Comparison with Ensembling of \coop.} 
In Appendix~\ref{sec:ensemble}, we conduct an experiment to determine if directly combining multiple \coop\ prompts can lead to performance improvement. The results demonstrate that combining 2, 4, or 6 \coop\ prompts does not effectively enhance performance and, at times, even deteriorates the performance. This indicates that our performance gains cannot be attributed to simple prompt ensembling.

\paragraph{Running Time. } 
In Appendix~\ref{sec:time-exps}, we conduct an experiment to compare the \coop, \cocoop, and \algo\ methods as shown in \autoref{tab:time-exps}. On the EuroSAT dataset, the runtime of \algo\ increased only slightly compared to \coop\ (14.1s vs. 34.1s), but it is significantly more efficient than \cocoop\ (62.0s), demonstrating the efficiency of the \algo\ algorithm.

\section{Related Work}

\paragraph{Few-shot Prompt Tuning. }
Prompt learning aims to formalize various NLP tasks to mask language modeling problems, which is similar to the pre-training of language models~\cite{devlin2018bert, radford2019language, clip} by adopting different prompt templates. 
The previous works~\cite{petroni2019language, radford2019language, brown2020language}
elaborately design human-crafted prompts, which is known as prompt engineering. 
Despite considerable progress in NLP, prompt learning remains underexplored in computer vision. Pretrained VLMs~\cite{jia2021scaling, clip} introduce hand-crafted prompts to perform zero-shot inference on the downstream tasks. 
However, designing specific prompts for various downstream tasks is inefficient and costly and several studies~\cite{shi24lft, shi24peft} peforms parameter-efficient fine-tuning to address this problem. 
CoOp~\cite{coop} makes prompt learnable via minimizing the classification loss on the target task, adopting the prompt tuning approach of NLP. However, CoOp decreases the zero-shot capability of VLMs. 
To fix the problem, CoCoOp~\cite{cocoop} introduces meta net to conditionally fine-tune prompts.
LFA~\cite{OualiBMT23} adopts a simple linear approach for vision-language alignment. 
VPT~\cite{derakhshani2022variational} attempts to learn a collection of continuous prompts to capture the variational visual representation. 
SHIP~\cite{ship} follows the paradigm of variational autoencoders to generate visual features according to the prompts via the generative method. 
ProDA~\cite{proda} proposes to learn the distribution of instance-specific prompts via variational inference. 
\citet{DingZYWXP24} explores the integration of OOD detection methods for VLMs and present meaningful observations. However, these studies do not consider the \setting\ evaluation setting.
Recent studies~\cite{Zhang0L24, ShuNHYGAX22} also make the attempts to perform prompt tuning on changing datastreams in a test-time adaptation manner~\cite{0007GJZL23, 20241667, zhao24sward}. These studies can be explored to address \setting\ problem in the furture studies.

\paragraph{Out-of-distribution Detection. }
Out-of-distribution detection refers to training the model on in-distribution (ID) dataset to classify OOD and ID samples. 
MSP~\cite{hendrycks2016baseline} takes the maximum softmax probability over ID categories as the score. 
RotPred~\cite{hendrycks2019using} includes an extra head to predict the rotation angle of rotated inputs in a self-supervised manner, and the rotation head together with the classification head is used for OOD detection. 
MCD~\cite{yu2019unsupervised} considers an ensemble of multiple classification heads and promotes the disagreement between each head’s prediction on OOD samples. 
StyleAugment~\cite{geirhos2018imagenet} applies style transfer to clean images to emphasize the shape bias over the texture bias.
STEP~\cite{ZhouGCLP21} focuses on exploring out-of-distribution detection within a semi-supervised setting~\cite{wei22robust, guo24robust, shi2024residual, jia2024lamda}. 
CIDER~\cite{ming2022exploit} regularizes the model’s hyperspherical space by increasing inter-class separability and intra-class compactness.
MixOE~\cite{zhang2023mixture} performs pixel-level mixing operations between ID and OOD samples and regularizes the model such that the prediction confidence smoothly decays as the input transitions from ID to OOD.
RegMixup~\cite{pinto2022using} trains the model with both clean images and mixed images obtained from the convex combination.
Recent studies~\cite{MingL24, DBLP:journals/corr/abs-2404-00323}, such as Clipn~\cite{wang2023clipn}, LoCoOp~\cite{locoop}, attempt to explore the capability of zero-shot and few-shot ood detection via VLMs respectively. However, while these studies primarily focus on OOD detection tasks, our research utilizes OOD detection to enhance the generalization of VLMs.

\section{Conclusion}
\label{sec:conclusion}
In this paper, we explore the \setting\ problem in detail and uncover that the base-to-new discriminability is crucial but often overlooked by existing methods and settings. 
We first introduce the \dept\ framework and demonstrate, through theoretical analysis, that incorporating OOD detection into prompt tuning can enhance the base-to-new discriminability and prevent degradation of new-class discriminability. 
Building upon \dept, we propose a novel prompt tuning approach called \algo\ that introduces new-class detectors and sub-classifiers to further enhance the discriminability of both the base and new classes. Experimental results validate our analysis of \dept\ and demonstrate the effectiveness of our \algo\ approach.

One limitation of our work is that we only take the initial step in combining OOD detection and prompt tuning. We believe there is potential for future researchers to explore. The other limitations are included in \autoref{sec:limitation}. 

\section*{Code Availability Statement}
The implementation code for this work is available at \url{https://github.com/WNJXYK/DeCoOp}. 

\vspace{-0.015in}
\section*{Acknowledgements}

This research was supported by National Science and Technology Major Project (2022ZD0114803) and the National Science Foundation of China (62306133, 62176118).

\vspace{-0.015in}
\section*{Impact Statement}
This paper aims to advance prompt tuning for vision-language models. 
The work carried out in this study has various potential societal implications. We firmly believe that the majority of these impacts are positive and it is unnecessary to explicitly emphasize any specific ones in this
paper. Additionally, we anticipate that the responsible utilization of technology will foster discourse concerning the best practices and regulations for implementing methods. 

\bibliography{ref}
\bibliographystyle{icml2024}


\newpage
\appendix
\onecolumn
\section{Proof of \autoref{thm: error}}
\label{sec: proof_of_error}
\begin{proof}
We first compute $H^{\cls}_{\zs}(\boldsymbol{x})$ and $H^{\ood}_{\zs}(\boldsymbol{x})$ for one specific instance $\boldsymbol{x}$. 
Recall that for an instance $\boldsymbol{x}$, we denote its ground-truth label space as $k$ (which always equals to $g(\boldsymbol{x})$) and its ground-truth label as $f(\boldsymbol{x})$. 
To facilitate the proof, we define additional label spaces:
\begin{equation}
    \mathcal{Y}_{i, j} = 
    \begin{cases}
        \{j\}, & j \in \mathcal{Y}_i, \\
        \emptyset, & otherwise, 
    \end{cases}
\end{equation}
and additional class vectors for $\boldsymbol{x}$: 
\begin{equation}
    \tilde{y}_{i, j} = 
    \begin{cases}
        1, & f(\boldsymbol{x}) = j \land f(\boldsymbol{x}) \in \mathcal{Y}_i, \\
        0, & otherwise. 
    \end{cases}
\end{equation}
Our computational results are presented as follows:
\begin{equation}
\begin{aligned}
H^{\cls}_{\zs}(\boldsymbol{x})
&= H \left (\tilde{\boldsymbol{y}}, \{P_{\zs}(y = j | y \in \mathcal{Y}_{k}, \boldsymbol{x})\}_{j=1}^C \right ) \\
& = H \left (\tilde{\boldsymbol{y}}, \{P_{\zs}(y \in \mathcal{Y}_{{k}, j} | y \in \mathcal{Y}_{k}, \boldsymbol{x})\}_{j=1}^C \right ) \\
&=  - \sum_{j=1}^C \tilde{y}_j \log{ P_{\zs}(y = j| y \in \mathcal{Y}_{k}, \boldsymbol{x}) } \\
&=  - \log{ P_{\zs}(y = f(\boldsymbol{x})  | y \in \mathcal{Y}_{k}, \boldsymbol{x}) },
\end{aligned}    
\end{equation}
and
\begin{equation}
\begin{aligned}
 H^{\ood}_{\zs}(\boldsymbol{x})
&= H \left (\tilde{\boldsymbol{k}}, \{P_{\zs}(y \in \mathcal{Y}_i| \boldsymbol{x})\}_{i=\{\text{b}, \text{n}\}} \right ) \\
&= - \sum_{i\in\{\text{b}, \text{n}\}} \tilde{k}_i \log{ P_{\zs} (y \in \mathcal{Y}_{i}| \boldsymbol{x})  } \\
&= - \log{ P_{\zs} (y \in \mathcal{Y}_{k} | \boldsymbol{x}) } . 
\end{aligned}    
\end{equation}

Then, we can bound $\mathbb{E}_{\boldsymbol{x}}\left [ H_{\zs}(\boldsymbol{x})\right ]$ as follows:
\begin{equation}
\begin{aligned}
\mathbb{E}_{\boldsymbol{x}}\left [ H_{\zs}(\boldsymbol{x}) \right ]
&= \mathbb{E}_{\boldsymbol{x}}\left [ H \left (\tilde{\boldsymbol{y}}, \{P_{\zs}(y = j| \boldsymbol{x})\}_{j=1}^C \right ) \right ]\\
&= \mathbb{E}_{\boldsymbol{x}}\left [ H \left (\tilde{\boldsymbol{y}}, \{P_{\zs}(y \in \mathcal{Y}_{k, j}| \boldsymbol{x})\}_{j=1}^C \right ) \right ]\\
&= \mathbb{E}_{\boldsymbol{x}}\left [ - \log{ P_{\zs}(y \in \mathcal{Y}_{k, f(\boldsymbol{x}) }| \boldsymbol{x}) } \right ]\\
&= \mathbb{E}_{\boldsymbol{x}}\left [ - \log{ P_{\zs}(y \in \mathcal{Y}_{k, f(\boldsymbol{x}) }| y \in \mathcal{Y}_{k}, \boldsymbol{x}) } - \log{ P_{\zs}(y \in \mathcal{Y}_{k} | \boldsymbol{x}) } \right ]\\
&= \mathbb{E}_{\boldsymbol{x}}\left [ H^{\cls}_{\zs}(\boldsymbol{x}) + H^{\ood}_{\zs}(\boldsymbol{x}) \right ]\\
&= \mathbb{E}_{\boldsymbol{x}}\left [ H^{\cls}_{\zs}(\boldsymbol{x}) \right ]
 + \mathbb{E}_{\boldsymbol{x}}\left [ H^{\ood}_{\zs}(\boldsymbol{x}) \right ] \\
&\leq \delta + \epsilon.
\end{aligned}
\end{equation}

Further, we can similarily compute $H^{\cls}_{\pt}(\boldsymbol{x})$ as follows:
\begin{equation}
\begin{aligned}
H^{\cls}_{\pt}(\boldsymbol{x})
&= H \left (\tilde{\boldsymbol{y}}, \{P_{\pt}(y = j | y \in \mathcal{Y}_{k}, \boldsymbol{x})\}_{j=1}^C \right ) \\
&= H \left (\tilde{\boldsymbol{y}}, \{P_{\pt}(y \in \mathcal{Y}_{k, j}| y \in \mathcal{Y}_{k}, \boldsymbol{x})\}_{j=1}^C \right ) \\
&= - \sum_{j=1}^C \tilde{y}_{k, j} \log{ P_{\pt}(y \in \mathcal{Y}_{k, j}| y \in \mathcal{Y}_{k}, \boldsymbol{x}) } \\
&= - \log{ P_{\pt}(y \in \mathcal{Y}_{k, f(\boldsymbol{x}) }| y \in \mathcal{Y}_{k}, \boldsymbol{x}) }.
\end{aligned}
\end{equation}

Finally, we can bound $\mathbb{E}_{\boldsymbol{x}}\left [ H_{\dept}(\boldsymbol{x})\right ]$ as follows:
\begin{equation}
\begin{aligned}
\mathbb{E}_{\boldsymbol{x}}\left [ H_{\dept}(\boldsymbol{x}) \right ]
=& ~ \mathbb{E}_{\boldsymbol{x}}\left [ H \left (\tilde{\boldsymbol{y}}, \{P_{\dept}(y = j| \boldsymbol{x})\}_{j=1}^C \right ) \right ] \\
=& ~ \mathbb{E}_{\boldsymbol{x}}\left [ H \left (\tilde{\boldsymbol{y}}, \{P_{\dept}(y \in \mathcal{Y}_{k, i}| \boldsymbol{x})\}_{i=1}^C \right ) \right ] \\
=& ~ \mathbb{E}_{\boldsymbol{x}}\left [ - \log{ P_{\dept}(y \in \mathcal{Y}_{k, i}| \boldsymbol{x}) } \right ] \\
=& ~ \mathbb{E}_{\boldsymbol{x}\land k=\text{b}} \left [ - \log{ P_{\dept}(y \in \mathcal{Y}_{k, i}| \boldsymbol{x}) } \right ] + \mathbb{E}_{\boldsymbol{x}\land k=\text{n}} \left [ - \log{ P_{\dept}(y \in \mathcal{Y}_{k, i}| \boldsymbol{x}) } \right ] \\
=& ~ \mathbb{E}_{\boldsymbol{x}\land k=\text{b}} \left [ - \log{ P_{\pt}(y \in \mathcal{Y}_{k, f(\boldsymbol{x}) }| y \in \mathcal{Y}_{k}, \boldsymbol{x}) } - \log{ P_{\zs}(y \in \mathcal{Y}_{k} | \boldsymbol{x}) } \right ] \\
&+  \mathbb{E}_{\boldsymbol{x}\land k=\text{n}} \left [ - \log{ P_{\zs}(y \in \mathcal{Y}_{k, f(\boldsymbol{x}) }| y \in \mathcal{Y}_{k}, \boldsymbol{x}) } - \log{ P_{\zs}(y \in \mathcal{Y}_{k} | \boldsymbol{x}) } \right ] \\
=& ~ \mathbb{E}_{\boldsymbol{x}\land k=\text{b}} \left [ H^{\cls}_{\pt}(\boldsymbol{x}) + H^{\ood}_{\zs}(\boldsymbol{x})  \right ] +  \mathbb{E}_{\boldsymbol{x}\land k=\text{n}} \left [ H^{\cls}_{\zs}(\boldsymbol{x}) + H^{\ood}_{\zs}(\boldsymbol{x}) \right ] \\
\leq & ~ \alpha \cdot (\delta - \Delta + \epsilon) + (1 - \alpha) \cdot (\delta + \epsilon) \\
\leq & ~ \delta + \epsilon - \alpha \cdot \Delta. 
\end{aligned}
\end{equation}
\end{proof}

\section{Additional Experimental Results}

\subsection{Detailed Results on ViT-B/32 Architecture}

To address the consistent performance of our proposal, we conduct experiments and compare our proposed DECOOP method, baseline methods, and SOTA prompting tuning methods on 11 datasets using
ViT-B/32 architectures. 
Each dataset is trained with random seeds from 1 to 5.
In terms of detailed performance on each dataset, our proposed method outperforms the comparison
methods on 9 out of 11 datasets, while achieving comparable performance on the remaining 2 datasets, showcasing its robustness to different pre-trained architectures. 

\label{sec:full-exps-vit32}
\begin{table*}[t]
    \centering
    \caption{Performance comparison between our proposed \algo\ method and comparison methods on 11 datasets using ViT-B/32 architecture. 
    The best performance is in bold. }
    \label{tab:full-exps-vit32}
    \vskip 0.15in
    \begin{center}
    \begin{small}
    \begin{sc}
    \resizebox{\textwidth}{!}{
    \begin{tabular}{l|cc|ll|ll|ll}
    \toprule
    \midrule
                    & \multicolumn{2}{c|}{Average}                            & \multicolumn{2}{c|}{ImageNet}                           & \multicolumn{2}{c|}{Caltech101}                         & \multicolumn{2}{c}{OxfordPets}                          \\
                    & \multicolumn{1}{c}{H}      & \multicolumn{1}{c|}{Acc.}  & \multicolumn{1}{c}{H}      & \multicolumn{1}{c|}{Acc.}  & \multicolumn{1}{c}{H}      & \multicolumn{1}{c|}{Acc.}  & \multicolumn{1}{c}{H}      & \multicolumn{1}{c}{Acc.}   \\ \midrule 
    Clip            & 67.13                      & 60.36                      & 65.69 $\pm$ 0.00           & 62.05 $\pm$ 0.00           & 93.78 $\pm$ 0.00           & 91.08 $\pm$ 0.00           & 91.30 $\pm$ 0.00           & 85.01 $\pm$ 0.00           \\ 
    Prompt Ens. & 67.76                      & 60.73                      & 66.91 $\pm$ 0.00           & 63.22 $\pm$ 0.00           & 94.06 $\pm$ 0.00           & 91.20 $\pm$ 0.00           & 89.73 $\pm$ 0.00           & 83.18 $\pm$ 0.00           \\ 
    CoOp            & 67.86                      & 61.03                      & 60.99 $\pm$ 0.09           & 57.61 $\pm$ 0.12           & 93.55 $\pm$ 0.76           & 91.09 $\pm$ 0.45           & 92.17 $\pm$ 0.77           & 85.21 $\pm$ 0.65           \\ 
    CoCoOp          & 70.77                      & 62.96                      & 67.74 $\pm$ 1.23           & 64.06 $\pm$ 1.39           & 93.78 $\pm$ 0.92           & 91.01 $\pm$ 0.87           & \textbf{94.05 $\pm$ 0.56}           & \textbf{87.84 $\pm$ 0.89}           \\ 
    Ship            & 69.25                      & 59.91                      & 61.72 $\pm$ 0.61           & 56.93 $\pm$ 1.26           & 93.35 $\pm$ 0.93           & 89.80 $\pm$ 0.83           & 92.19 $\pm$ 1.47           & 81.22 $\pm$ 1.03           \\ 
    DeCoOp(Ours)        & \textbf{72.51}                      & \textbf{65.75}                      & \textbf{68.07 $\pm$ 0.06}           & \textbf{64.49 $\pm$ 0.04}           & \textbf{95.56 $\pm$ 0.22}           & \textbf{93.36 $\pm$ 0.48}           & 93.13 $\pm$ 0.50          & 86.25 $\pm$ 0.96         \\ 
    \midrule
    \midrule
                    & \multicolumn{2}{c|}{StandfordCars}                      & \multicolumn{2}{c|}{Flowers102}                         & \multicolumn{2}{c|}{Food101}                            & \multicolumn{2}{c}{FGVCAircraft}                        \\
                    & \multicolumn{1}{c}{H}      & \multicolumn{1}{c|}{Acc.}  & \multicolumn{1}{c}{H}      & \multicolumn{1}{c|}{Acc.}  & \multicolumn{1}{c}{H}      & \multicolumn{1}{c|}{Acc.}  & \multicolumn{1}{c}{H}      & \multicolumn{1}{c}{Acc.}   \\ \midrule 
    Clip            & 65.14 $\pm$ 0.00           & 60.39 $\pm$ 0.00           & 70.50 $\pm$ 0.00           & 64.27 $\pm$ 0.00           & 85.10 $\pm$ 0.00           & 79.16 $\pm$ 0.00           & 23.62 $\pm$ 0.00           & 18.30 $\pm$ 0.00           \\ 
    Prompt Ens. & 64.67 $\pm$ 0.00           & 59.82 $\pm$ 0.00           & 68.60 $\pm$ 0.00           & 63.30 $\pm$ 0.00           & 85.55 $\pm$ 0.00           & 79.59 $\pm$ 0.00           & 23.45 $\pm$ 0.00           & 18.30 $\pm$ 0.00           \\ 
    CoOp            & 62.33 $\pm$ 1.21           & 56.95 $\pm$ 1.37           & 71.13 $\pm$ 1.95           & 65.25 $\pm$ 1.43           & 81.55 $\pm$ 0.91           & 74.32 $\pm$ 1.17           & 23.15 $\pm$ 1.71           & 18.88 $\pm$ 0.85           \\ 
    CoCoOp          & 65.48 $\pm$ 0.66           & 60.27 $\pm$ 0.84           & 74.46 $\pm$ 1.10           & 65.86 $\pm$ 1.53           & \textbf{86.11 $\pm$ 0.29}           & \textbf{80.09 $\pm$ 0.40}           & 21.68 $\pm$ 5.89           & 15.28 $\pm$ 4.87           \\ 
    Ship            & 64.38 $\pm$ 0.81           & 56.22 $\pm$ 1.00           & 70.41 $\pm$ 1.72           & 62.41 $\pm$ 1.88           & 81.76 $\pm$ 0.90           & 72.14 $\pm$ 1.43           & 19.34 $\pm$ 2.64           & 19.00 $\pm$ 0.98           \\ 
    DeCoOp(Ours)        & \textbf{67.45 $\pm$ 0.15}           & \textbf{62.55 $\pm$ 0.23}           & \textbf{79.06 $\pm$ 0.43}           & \textbf{72.84 $\pm$ 0.77}           & 86.04 $\pm$ 0.10          & 79.98 $\pm$ 0.11        & \textbf{25.58 $\pm$ 0.33}           & \textbf{20.03 $\pm$ 0.16}           \\ 
    \midrule
    \midrule
                    & \multicolumn{2}{c|}{SUN397}                             & \multicolumn{2}{c|}{DTD}                                & \multicolumn{2}{c|}{EuroSAT}                            & \multicolumn{2}{c}{UCF101}                              \\
                    & \multicolumn{1}{c}{H}      & \multicolumn{1}{c|}{Acc.}  & \multicolumn{1}{c}{H}      & \multicolumn{1}{c|}{Acc.}  & \multicolumn{1}{c}{H}      & \multicolumn{1}{c|}{Acc.}  & \multicolumn{1}{c}{H}      & \multicolumn{1}{c}{Acc.}   \\ \midrule 
    Clip            & 71.35 $\pm$ 0.00           & 61.99 $\pm$ 0.00           & 53.60 $\pm$ 0.00           & 42.85 $\pm$ 0.00           & 50.81 $\pm$ 0.00           & 38.17 $\pm$ 0.00           & 67.56 $\pm$ 0.00           & 60.67 $\pm$ 0.00           \\ 
    Prompt Ens. & 73.27 $\pm$ 0.00           & 63.74 $\pm$ 0.00           & 53.81 $\pm$ 0.00           & 43.44 $\pm$ 0.00           & 56.90 $\pm$ 0.00           & 40.75 $\pm$ 0.00           & 68.39 $\pm$ 0.00           & 61.49 $\pm$ 0.00           \\ 
    CoOp            & 69.48 $\pm$ 1.01           & 59.89 $\pm$ 0.85           & 57.52 $\pm$ 1.82           & 48.90 $\pm$ 1.23           & 67.46 $\pm$ 7.70           & 51.07 $\pm$ 8.05           & 67.11 $\pm$ 3.56           & 62.12 $\pm$ 2.48           \\ 
    CoCoOp          & 75.51 $\pm$ 0.37           & 65.96 $\pm$ 0.45           & 59.57 $\pm$ 2.21           & 47.08 $\pm$ 1.30           & 66.98 $\pm$ 8.67           & 49.19 $\pm$ 5.78           & 73.17 $\pm$ 1.24           & 65.98 $\pm$ 1.06           \\ 
    Ship            & 70.33 $\pm$ 0.63           & 58.86 $\pm$ 0.71           & 57.22 $\pm$ 3.14           & 45.91 $\pm$ 1.07           & \textbf{77.74 $\pm$ 3.74}           & 50.23 $\pm$ 1.92           & 73.27 $\pm$ 1.21           & 66.31 $\pm$ 0.72           \\ 
    DeCoOp(Ours)        & \textbf{75.87 $\pm$ 0.14}           & \textbf{66.59 $\pm$ 0.19}           & \textbf{60.61 $\pm$ 0.48}           & \textbf{50.39 $\pm$ 0.40}           & 72.35 $\pm$ 2.42      & \textbf{58.93 $\pm$ 2.62}           & \textbf{73.87 $\pm$ 0.36}           & \textbf{67.83 $\pm$ 0.81}           \\ 
    \midrule
    \bottomrule
    \end{tabular}
    }
    \end{sc}
    \end{small}
    \end{center}
    \vskip -0.1in
\end{table*}

\subsection{Detailed AUROC Results}
\label{sec:full-exps-auroc}

The full experimental results of \autoref{tab:auroc-exps} are presented in \autoref{tab:full-auroc-exps}. Our \algo\ approach achieves the best base-to-new discriminability among all comparison methods. 

\begin{table}[t]
    \centering
    \caption{AUROC performance is compared with \Clip, Prompt Ensemble, \coop, \cocoop, \ship \ and our proposed \algo. The results demonstrate that our proposal enhances base-to-new discriminability.}
    \label{tab:full-auroc-exps}
    \vskip 0.15in
    \begin{center}
    \begin{small}
    \begin{sc}
    \resizebox{\linewidth}{!}{
    \begin{tabular}{l|cccccc}
    \toprule
    \midrule
    dataset & \Clip & \Pens & \coop & \cocoop & \ship & \algo(Ours)  \\
    \midrule
    ImageNet        & 88.34 $\pm$ 0.00 & 89.79 $\pm$ 0.00 & 77.14 $\pm$ 1.62 & 88.05 $\pm$ 1.22 & 84.71 $\pm$ 1.62 & \textbf{97.48 $\pm$ 0.03} \\ 
    Caltech101      & 97.03 $\pm$ 0.00 & 97.09 $\pm$ 0.00 & 94.53 $\pm$ 0.87 & 95.71 $\pm$ 0.50 & 96.94 $\pm$ 0.79 & \textbf{99.58 $\pm$ 0.03} \\ 
    OxfordPets      & 92.66 $\pm$ 0.00 & 92.21 $\pm$ 0.00 & 91.06 $\pm$ 1.00 & 91.15 $\pm$ 0.95 & 93.30 $\pm$ 1.23 & \textbf{98.12 $\pm$ 0.24} \\ 
    StanfordCars    & 86.24 $\pm$ 0.00 & 87.46 $\pm$ 0.00 & 78.25 $\pm$ 2.00 & 83.00 $\pm$ 2.24 & 87.23 $\pm$ 1.16 & \textbf{97.63 $\pm$ 0.02} \\ 
    Flowers102      & 84.92 $\pm$ 0.00 & 87.78 $\pm$ 0.00 & 78.06 $\pm$ 1.82 & 79.63 $\pm$ 2.20 & 84.84 $\pm$ 1.41 & \textbf{95.75 $\pm$ 0.18} \\ 
    Food101         & 89.88 $\pm$ 0.00 & 90.26 $\pm$ 0.00 & 87.53 $\pm$ 1.20 & 88.19 $\pm$ 1.07 & 89.92 $\pm$ 1.00 & \textbf{97.59 $\pm$ 0.04} \\ 
    FGVCAircraft    & 75.08 $\pm$ 0.00 & 75.86 $\pm$ 0.00 & 75.25 $\pm$ 1.36 & 69.00 $\pm$ 7.91 & 75.78 $\pm$ 1.65 & \textbf{84.06 $\pm$ 0.26} \\ 
    SUN397          & 72.46 $\pm$ 0.00 & 75.29 $\pm$ 0.00 & 70.29 $\pm$ 1.47 & 73.75 $\pm$ 1.11 & 74.78 $\pm$ 1.14 & \textbf{90.21 $\pm$ 0.10} \\ 
    DTD             & 62.29 $\pm$ 0.00 & 61.10 $\pm$ 0.00 & 56.78 $\pm$ 1.93 & 60.65 $\pm$ 0.94 & 60.66 $\pm$ 1.22 & \textbf{75.47 $\pm$ 1.02} \\ 
    EuroSAT         & 56.40 $\pm$ 0.00 & 57.74 $\pm$ 0.00 & 52.26 $\pm$ 8.68 & 57.74 $\pm$ 2.49 & 59.32 $\pm$ 6.31 & \textbf{77.78 $\pm$ 3.85} \\ 
    UCF101          & 82.03 $\pm$ 0.00 & 83.56 $\pm$ 0.00 & 72.72 $\pm$ 2.21 & 79.03 $\pm$ 1.52 & 80.35 $\pm$ 1.99 & \textbf{93.56 $\pm$ 0.62} \\ \hline
    Average         & 80.67 & 81.65 & 75.81 & 78.72 & 80.71 & \textbf{91.57} \\ 
    \midrule
    \bottomrule  
    \end{tabular}}
    \end{sc}
    \end{small}
    \end{center}
    \vskip -0.1in
\end{table}

\subsection{Detailed ROC Curves}
\label{sec:full-exps-roc}
\begin{figure}[t]
    \vskip 0.2in
    \begin{center}
    \centerline{\includegraphics[width=\columnwidth]{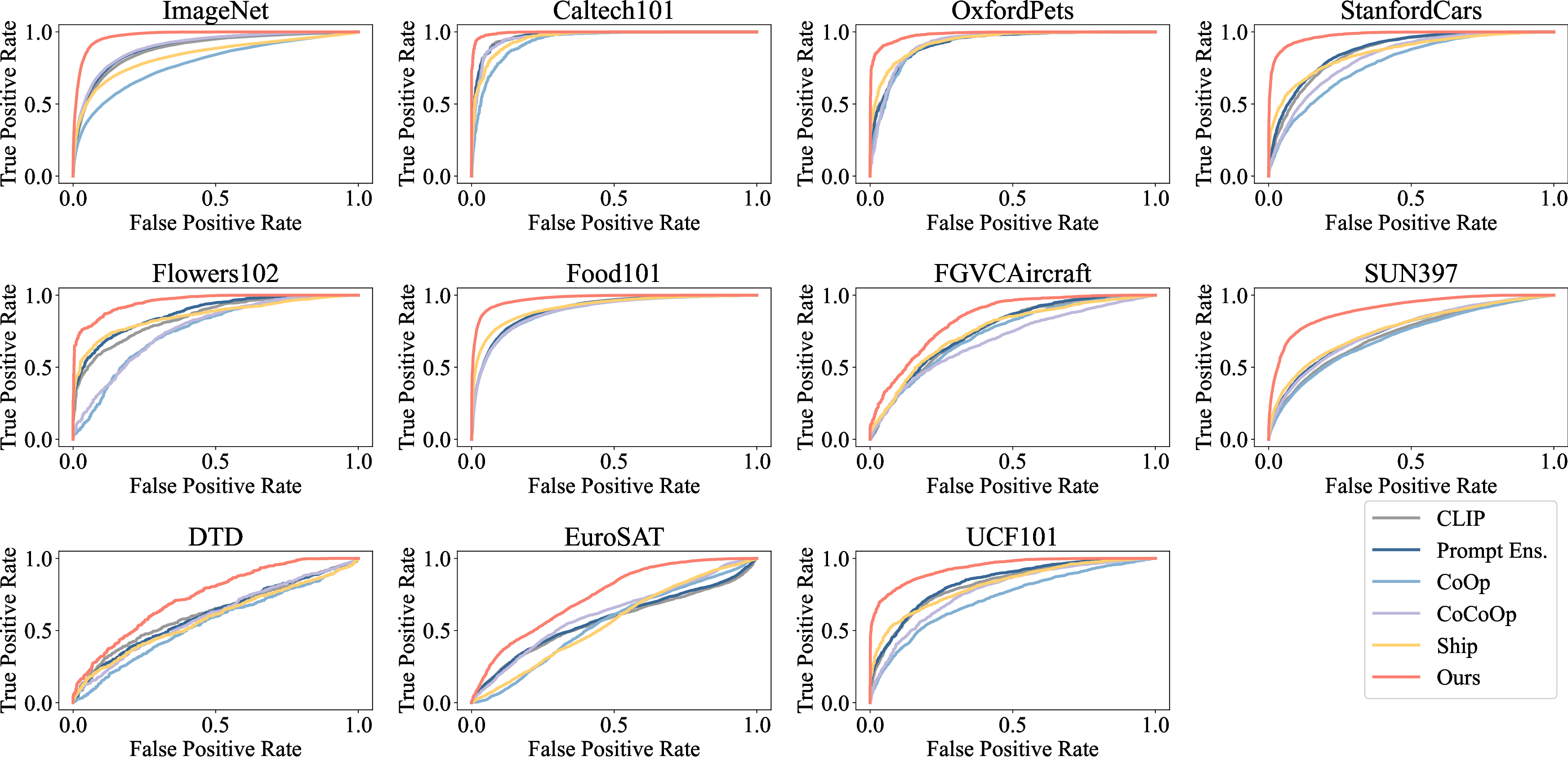}}
    \caption{The roc curve for detecting new classes of each method on 11 datasets.}
    \label{fig: full-auroc}
    \end{center}
    \vskip -0.2in
\end{figure}

To evaluate whether our proposal can improve the performance for detecting, we conduct experiments on 11 datasets using ViT-B/16 architecture. 
Each curve is drawn using our experiment results with random seeds to 1. 
For each method, we adopt the maximum softmax probability over new classes as the detecting score for drawing the curve. 
The results in \autoref{fig: full-auroc} show that our proposal can achieve the best detection performance. 

\subsection{Correlation between $\mathcal{M}_O$ and Performance}
\label{sec:correlation}

\begin{figure}[t]
    \vskip 0.2in
    \begin{center}
    \includegraphics[width=0.5\linewidth]{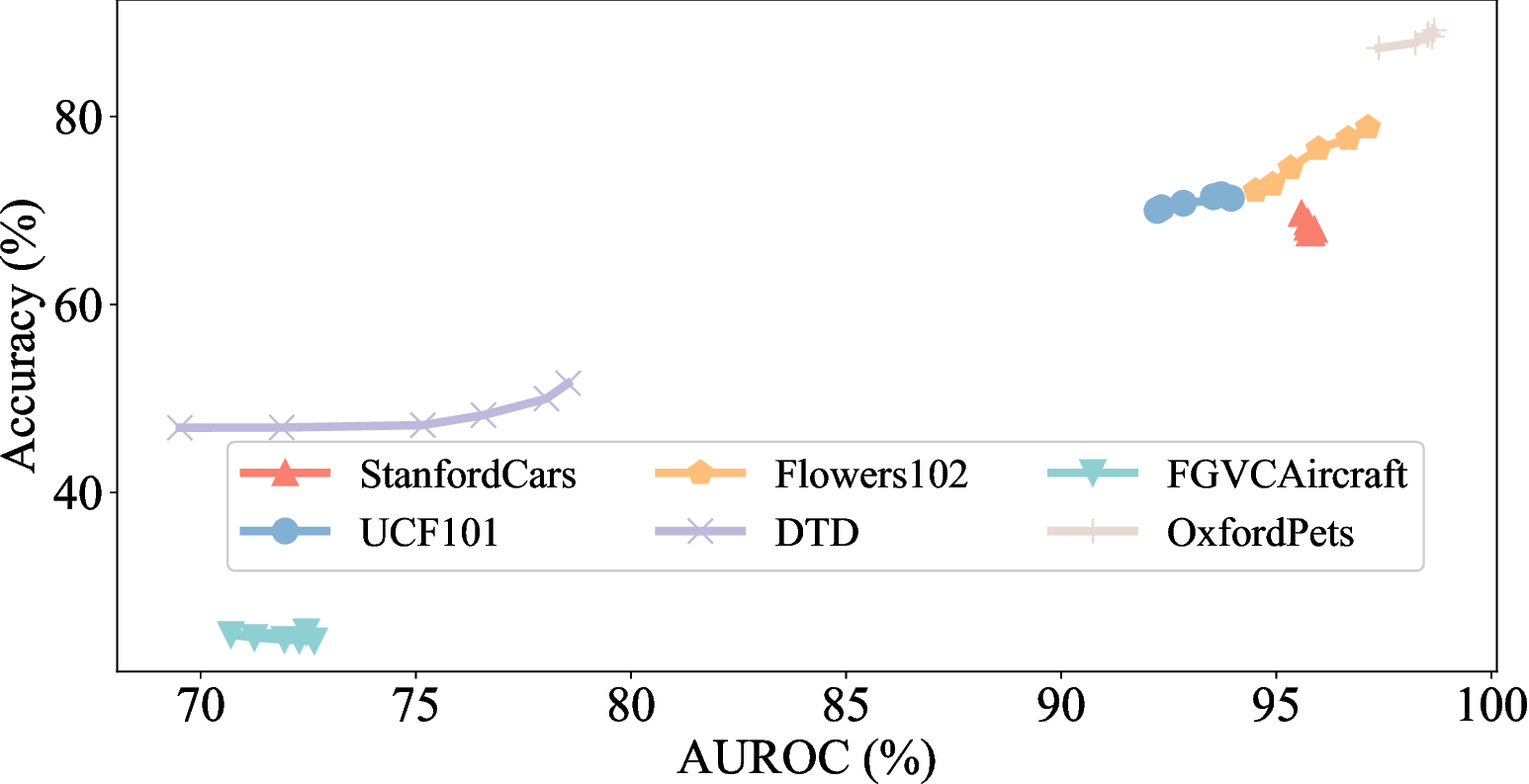}
    \caption{Correlation between performance of $\mathcal{M}_O$ and accuracy.}
    \label{fig: correlation}
    \end{center}
    \vskip -0.2in
\end{figure}

The objective of the \algo\ approach is to enhance the base-to-new discriminability through the $\mathcal{M}_O$, leading to improved performance. Hence, a key question arises: does a better $\mathcal{M}_O$ result in enhanced performance? To investigate this, we employ different new-class detectors with varying AUROC values for training and evaluate the performance as shown in \autoref{fig: correlation}.  
This figure illustrates the correlation between the AUROC of the new-class detector and the performance metric. 
The results indicate a positive correlation between these two variables, validating our claim and aligning with our research objective.

\subsection{Ablation Study}
\label{sec:ablation}

\begin{table}[t]
    \centering
    \caption{
    Ablation study. We report average performance across 11 datasets was conducted among baselines, \coop, \dept, \algo\ approaches, utilizing ViT-B/16 and ViT-B/32 architectures. 
    The best performance is in bold. The second-best performance is underlined. }
    \label{tab:ablation-exps}
    \begin{center}
    \begin{small}
    \begin{sc}
    \begin{tabular}{l|cc|cc}
    \toprule
    \midrule
    \multirow{2}{*}{Method} & \multicolumn{2}{c}{ViT-B/16} & \multicolumn{2}{c}{ViT-B/32}\\ \cline{2-5} 
                            & H    & Accuracy  & H & Accuracy   \\ \midrule
    \Clip & 70.84 & 63.92  & 67.13 & 60.36  \\ 
    \Pens & 71.65 & 65.39  & 67.76 & 60.73  \\ 
    \coop & 72.14 & 65.57  & 67.86 & 61.03  \\ \hline
    \dept & \underline{74.82} & \underline{68.03}  & \underline{69.96} & \underline{62.92}  \\ 
    \algo & \textbf{76.13} & \textbf{69.69}  & \textbf{72.51} & \textbf{65.75}  \\ 
    \midrule
    \bottomrule  
    \end{tabular}
    \end{sc}
    \end{small}
    \end{center}
    \vskip -0.1in
\end{table}

We conduct ablation studies to validate the effectiveness of each component of our proposed \algo\ approach in \autoref{tab:ablation-exps}. 
In this paper, we first propose a novel prompt tuning framework \dept\ to introduce OOD detection into prompt tuning. Then, two advanced modules are integrated into \dept\ framework to form our \algo\ approach. 
As the our two modules cannot be separated to perform classification, we compare baseline methods, \dept\ framework, and our proposed \algo\ appraoch. 
The results show that \dept\ framework enhances the base-to-new discriminability and prevents performance degradation of new classes, thereby outperforming \Clip, \Pens, and \coop\ methods.
Further, our proposed \algo\ approach achieves the best performance among all methods, demonstrating it additionally enhances the base-class and new-class discriminability.

\subsection{Simple Ensembling of \coop\ Method}
\label{sec:ensemble}

We also conduct experiments to evaluate whether directly ensemble multiple \coop\ learners can achieve similar performance. The results, shown in \autoref{tab:kcoop-exps}, indicate that the ensemble of multiple \coop\ prompts does not yield significantly better performance compared to the \coop\ method. These results prove that the performance gain does not derive from simple prompt ensembling.

\begin{table}[t]
    \centering
    \caption{Performance comparison between our proposed method and the ensemble of multiple \coop\ prompts is conducted. The results demonstrate that directly combining multiple \coop\ learners does not yield significantly better performance compared to the \coop\ method. Moreover, our proposed algorithm outperforms other methods.}
    \label{tab:kcoop-exps}
    \vskip 0.15in
    \begin{center}
    \begin{small}
    \begin{sc}
    \begin{tabular}{l|cccc}
    \toprule
    \midrule
    method & Flowers102 & DTD & Caltech101 & StandfordCars  \\
    \midrule
    \coop                & 72.11   & 48.18   & 93.24   & 63.81   \\ 
    \coop $\times 2$     & 71.62   & 50.08   & 93.31   & 64.20   \\ 
    \coop $\times 4$     & 73.12   & 49.99   & 92.89   & 64.32   \\ 
    \coop $\times 6$     & 71.89   & 49.69   & 92.87   & 65.03   \\  \hline
    \algo                & \textbf{78.61}   & \textbf{51.44}   & \textbf{94.50}   & \textbf{69.64}   \\ 
    \midrule
    \bottomrule  
    \end{tabular}
    \end{sc}
    \end{small}
    \end{center}
    \vskip -0.1in
\end{table}

\subsection{Evaluation Time}
\label{sec:time-exps}

Our \algo\ approach adopts multiple prompts to detect OOD, so it may take more time. 
We compared the running time taken by \coop, \cocoop, and \algo\ methods when evaluating the testing set of two datasets in \autoref{tab:time-exps}. 
The results show that the running time of the \algo\  is not significantly longer than the \coop\ method since the computation can be performed in parallel. However, our \algo\  approach runs in two stages (i.e., OOD detection and classification stages), therefore, the running time will be approximately double compared to the \coop\ method. 
However, the running time of \cocoop\ rises significantly as the number of categories increases, where demonstates our \algo\ is efficient.

\begin{minipage}{\textwidth}
 \begin{minipage}[t]{0.48\textwidth}
  \centering
     \makeatletter\def\@captype{table}\makeatother\caption{Evaluation running time of \coop, \cocoop, and \algo\ methods.}
    \label{tab:time-exps}
    \vskip 0.15in
    \begin{center}
    \begin{small}
    \begin{sc}
    \resizebox{\linewidth}{!}{
    \begin{tabular}{l|r|rrr}
    \toprule
    \midrule
    Datasets & \#Classes & \coop & \cocoop & \algo  \\
    \midrule
    EuroSAT & 10 & 14.1S & 62.0S & 34.1S \\
    Food101 & 101 & 50.5S & 711.7S & 131.5S \\
    \midrule
    \bottomrule  
    \end{tabular}}
    \end{sc}
    \end{small}
    \end{center}
    \vskip -0.1in
  \end{minipage}
  \hfill
  \begin{minipage}[t]{0.48\textwidth}
   \centering
        \makeatletter\def\@captype{table}\makeatother\caption{Comparison with weight interpolating methods. }
    \label{tab:weight-exps}
    \vskip 0.15in
    \begin{center}
    \begin{small}
    \begin{sc}
    \resizebox{\linewidth}{!}{
    \begin{tabular}{l|cc}
    \toprule
    \midrule
     & New Acc.	 & Accuracy\\
    \midrule
    CLIP	& 65.48	& 63.92 \\
    CoOp	& 57.75	& 65.58 \\
    RFT~\cite{WortsmanIKLKRLH22}	& 65.34	& 69.26 \\
    DeCoOp	& \textbf{66.54}	& \textbf{69.69} \\
    \midrule
    \bottomrule  
    \end{tabular}}
    \end{sc}
    \end{small}
    \end{center}
    \vskip -0.1in
   \end{minipage}
\end{minipage}

\subsection{Comparison with Weight Interpolating Methods}

Existing studies~\cite{WortsmanIKLKRLH22, IlharcoWGSHKFS22} observe that interpolating weights for tuned and original vision-language models can improve the generalization capacity. In the context of prompt tuning, we can interpolate weights of the tuned prompt and original prompt. We report the average results on all datasets using ViT-B/16 architecture in \autoref{tab:weight-exps}. 
The results show that interpolating weights can give better performance compared to both the original model and the tuned model, which aligns with the conclusion of existing studies. Our \algo\ outperforms other methods, demonstrating its effectiveness. Note that Weight Interpolating Methods and \algo\ have studied the different stages in fine-tuning, therefore, the combination of both to further enhance performance can be a direction for future research.

\section{Limitation and Future Work}
\label{sec:limitation}

Our paper proposes the integration of OOD detection into prompt tuning to prevent performance degradation on new classes. 
In addition to the content discussed at the end of Section~\ref{sec:conclusion}, one limitation of our approach is the potentially increased time consumption due to the adoption of a two-stage classification process. 
Integrating knowledge~\cite{yangshortcut, YangSTLDZ24} into the prompt tuning to achieve the better generalization is also a future direction to explore. 
Experiments detailed in Appendix~\ref{sec:time-exps} demonstrate that our method's running time is shorter than some existing methods (e.g., \cocoop), proving that the running time of our proposal is within an acceptable range. One possible solution is to integrate the two-stage classification into prompt training through the utilization of advanced training strategies, which can be explored as potential research directions in the future. 

\end{document}